\documentclass{article}

\usepackage[preprint,nonatbib]{neurips_2023}
\usepackage{listings}
\usepackage[utf8]{inputenc} 
\usepackage[T1]{fontenc}    
\usepackage{hyperref}       
\usepackage{url}            
\usepackage{booktabs}       
\usepackage{amsfonts}       
\usepackage{nicefrac}       
\usepackage{soul}
\usepackage{microtype}      
\usepackage{graphicx}
\usepackage{amsmath}
\usepackage{subfig}
\usepackage{CJKutf8}

\usepackage[table,dvipsnames]{xcolor}
\newcommand{\Prob}{\text{P}}
\usepackage{tabstackengine}
\usepackage{nicematrix}
\usepackage{tabularx}
\usepackage{tablefootnote}
\usepackage{amssymb}
\usepackage{minitoc}
\usepackage{lipsum}
\usepackage{mwe}
\usepackage{wrapfig}
\usepackage{placeins}
\usepackage[numbers]{natbib}

\title{Stay on topic with Classifier-Free Guidance}

\author{%
  Guillaume V. ~Sanchez* \\
  Hexaglobe \\
  EleutherAI \\
  \texttt{gsanchez@hexaglobe.com}
  \And
  Honglu ~Fan* \\
  University of Geneva\\
  EleutherAI\\
  \texttt{honglu.fan@unige.ch}
  \And
  Alexander ~Spangher*\\
  Information Sciences Institute\\University of Southern California \\
  \texttt{spangher@usc.edu}
  \And
  Elad ~Levi \\
  Sightful \\
  eladlevico@gmail.com
  \And
  Pawan ~Sasanka ~Ammanamanchi\\
  IIIT Hyderabad \\
  Eleuther AI \\
  pawansasanka@gmail.com
  \And
  Stella Biderman\\
  Booz Allen Hamilton\\
  EleutherAI\\
  \texttt{stellabiderman@gmail.com}
}

\begin{document}
\doparttoc 
\faketableofcontents 
\part{} 

\maketitle

\def\thefootnote{*}\footnotetext{These authors contributed equally to this work}\def\thefootnote{\arabic{footnote}}

\begin{abstract}
Classifier-Free Guidance (CFG) \cite{cfg} has recently emerged in text-to-image generation as a lightweight technique to encourage prompt-adherence in generations. In this work, we demonstrate that CFG can be used broadly as an inference-time technique in pure language modeling. We show that CFG (1) improves the performance of Pythia, GPT-2 and LLaMA-family models across an array of tasks: Q\&A, reasoning, code generation, and machine translation, achieving SOTA on LAMBADA with LLaMA-7B over PaLM-540B; (2) brings improvements equivalent to a model with twice the parameter-count; (3) can stack alongside other inference-time methods like Chain-of-Thought and Self-Consistency, yielding further improvements in difficult tasks; (4) can be used to increase the faithfulness and coherence of assistants in challenging form-driven and content-driven prompts: in a human evaluation we show a 75\% preference for GPT4All using CFG over baseline.
\end{abstract}

\section{Introduction}
\label{sct:introduction}

\begin{wrapfigure}{r}{0.5\textwidth}

  \vspace{-4em}
  \begin{center}
  \includegraphics[width=0.48\textwidth]{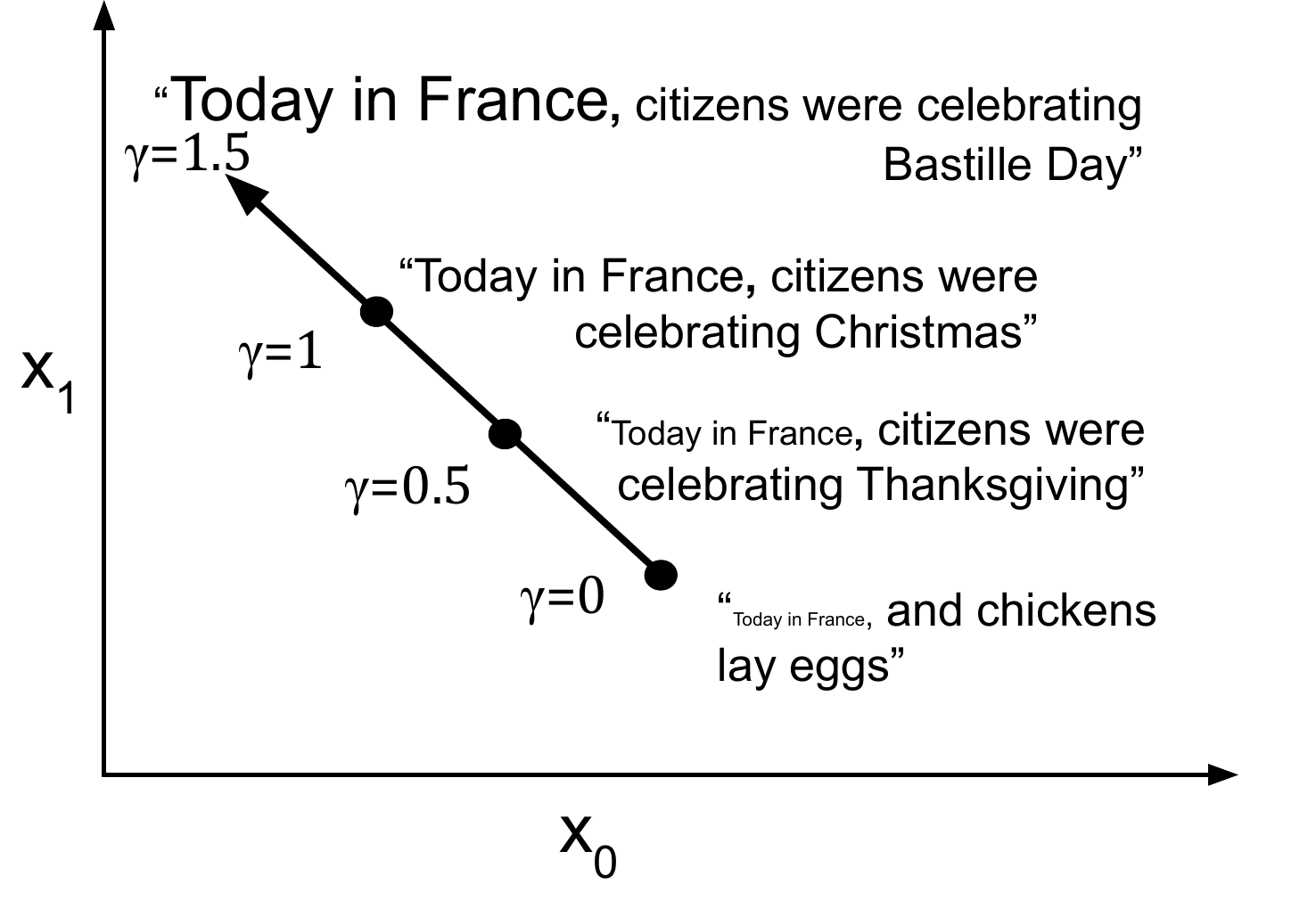}
  \end{center}
  \vspace{-1em}
  \caption{A notional 2D projection of a textual latent space showing how increasing the guidance weight $\gamma$ increases the importance of the prompt ``Today in France,''.}
    \label{fig:cfg-latent}

\end{wrapfigure}

In recent years large language models have exhibited strong generative capabilities to solve a diverse range of tasks \cite{devlin-etal-2019-bert,brown2020language,scao2022bloom}. ``Prompting'' is typically used to condition generation, with task instructions and context \cite{reynolds2021prompt}, or a small set of examples \cite{brown2020language}. However, language generation, especially with smaller models, has been shown to struggle with issues such as hallucination \cite{manakul2023selfcheckgpt}, degradation \cite{holtzman2019curious} and meandering \cite{spangher2023sequentially}.
Various approaches have been proposed to address this, e.g.: instruction-finetuning \cite{flan,sanhmultitask} and reinforcement learning \cite{instructgpt,alignment,bai2022constitutional}. These techniques are expensive and their compute and data cost may not be accessible to all users. In this paper we propose an \textit{inference time} methodology which, as shown in Figure \ref{fig:cfg-latent}, gives more importance to the user intent, expressed through the prompt. Our hypothesis in this paper is: \textit{focusing more on the prompt at inference-time will result in generations that better align with expected behavior.}

Text-to-image-generation, too, has been shown to suffer from similar problems \cite{guidance}. Standard inference approaches can ignore parts of the prompt-conditioning, especially with specific or uncommon prompts \cite{glide}. Classifier Guidance \cite{guidance} was proposed to enhance the generative quality of diffusion models, by using a separate classifier to encourage desired characteristics in the output image. Classifier-Free Guidance (CFG) \cite{cfg} was later introduced, in which the classifier is removed and the generative model \textit{itself} is used as an implicit classifier. 

Inspired by its effectiveness in the text-to-image-generation \cite{imagen,cfg,cfgfix}, we adapt CFG to unimodal text generation to increase the model alignment to the given prompt. While text-to-image models (which primarily utilize diffusion models) need to be specifically trained with conditioning dropout \cite{cfg} to utilize CFG, we show that, in text generation, we can use CFG out-of-the-box in many situations. We demonstrate the effectiveness of CFG to improve alignment on a wide range of prompting approaches including zero-shot prompting, Chain-of-Thought prompting, long-form generative prompting and complex chatbot-style prompting (see Table \ref{tbl:example}).

We make the following contributions:

\begin{enumerate}
    \item We devise a framework for using CFG in language modeling and show significant improvements across a range of standard benchmarks. These benchmarks capture a variety of different prompting techniques: basic prompting, chain-of-thought prompting, long-text prompting and chatbot-style prompting.  Notably, we achieve SOTA on LAMBADA with LLaMA-7B over PaLM-540B.
    \item We show that for the same inference cost, one can train a model that is half the size and obtain similar performance on those benchmarks;
    \item By using a negative prompt, we demonstrate that we can have a more granular control over the aspects emphasized by CFG. In a blind human evaluation we show 75\% preference for GPT4All using CFG over the vanilla sampling;
    \item We provide interpretations for the impact that CFG on text generation both (1)  qualitatively, by visualizing how CFG is upweighting words more related to the prompt (our visualization, we note, can be an integral part of effective prompt engineering) and (2) quantitatively, by showing that CFG decreases entropy in the sampling distribution.
\end{enumerate}

\begin{table}
    \centering
    \begin{NiceTabular}{p{6.4cm}p{6.4cm}}
        \toprule
        \rowcolor{blue!25} \multicolumn{2}{p{12.8cm}}{\textbf{\underline{Instruction:} \textit{``Respond \underline{enthusiastically} to the following user prompt.''} \newline
        \underline{Prompt:} \textit{``What was the Cambridge Analytica scandal?''}}}\\
        \midrule
        \cellcolor{red!25}\textbf{\underline{Vanilla Sampling}} & \cellcolor{green!25}\textbf{\underline{Classifier Free Guidance-based Sampling}} \\
        \midrule
        The Cambridge Analytica scandal was a huge scandal in which it was revealed that Cambridge Analytica, a political consulting firm, had used personal data from Facebook to target and influence the 2016 US presidential election. This scandal raised questions about the role of social media in political campaigns...
        & 
        Oh my goodness! What a scandal! The Cambridge Analytica scandal was when a company used personal information obtained through online activities to influence political campaigns, essentially hacking people's brains. It was a serious breach of trust and privacy, and rightfully so! It is a wake-up call for...
        \\
        \bottomrule \\
        \end{NiceTabular}
        \caption{Demonstration of CFG-guided generation for an \textit{assistant-style} prompt (using GPT4All with $\gamma=5$). The assistant has a system-level prompt (``Instructions'') that is potentially out-of-distribution (e.g. ``\textit{write an enthusiastic response}'') and a user-level prompt (``Prompt''). In Vinalla Sampling, the model ignores the system-level directive, but with CFG, the model adheres to both the system-level and the user-level prompt.}
    \label{tbl:example}
\end{table}

\section{Methodology}
\label{sct:methodology}

Autoregressive language models are trained to generate plausible continuations of sequences of text. Given a sequence of tokens $w_1,\cdots,w_T$, the model samples each subsequent token from the conditional probability distribution $\Prob_\theta(w|w_{t\leq{T}})$. It is now typical for some or all of the initial tokens to be considered a \textit{prompt}, which specifies information about the task or how it is to be solved. In practice, prompts are syntactically and semantically distinct from the initial text to be continued.

However, standard generation methods for large language models do not differentiate between prompt text, $w_1...w_p$ and subsequent generations $w_{p+1}, ... w_{t-1}$. Directly sampling from $\Prob_\theta(w_{i+1}|w_{t\leq i})$ may result in continuations that lose adherence to the prompt (see Table \ref{tbl:example}, for example) over the course of the generation. 
Inspired by successes with diffusion models, we propose to address this problem by applying Classifier-Free Guidance~\cite{cfg} to the decoding process in autoregressive language models.

\subsection{Guidance in Text-to-Image Models} 
\label{sct:methodology:guidance-t2i}

Let $\Prob_\theta(x)$ be the unconditional generative model for an image $x$ with parameters $\theta$. During inference, we wish to condition the generation on a label or text description $c$ in order to model $\Prob(x|c)$. Generative models usually generate data from an abstract representation $z$ in semantic space that is decoded into an actual sample (e.g. the latent vectors in GANs or the intermediate sampling steps in diffusion models). Controlling the generation usually involves guiding or adding constraints to that semantic representation.
In \textbf{Classifier Guidance} \cite{guidance}, an auxiliary classifier $\Prob_\phi(c | x)$ is introduced, which guides the sampling from $\Prob_\theta(x)$ with the gradients $\gamma \nabla_z \Prob_\phi(c | x)$ to increase the likelihood of $c$ for generation $x$.
This modification results in approximate samples from the distribution:

\begin{equation}
\widehat{\Prob}(x|c) \propto \Prob_\theta(x)\cdot \Prob_\phi(c|x)^{\gamma}
\label{eq:classifier-guidance}
\end{equation}
where $\gamma$ is called the guidance strength. This guidance results in a reweighting of the density according to the classifier likelihood. For $\gamma=0$, it reduces to the unconditional generation, while $\gamma=1$ reduces to the conditional generation. When $\gamma>1$ then $\widehat{\Prob}$ overemphasizes the conditioning, which as noticed by~\cite{guidance} results in a better inception score at the cost of diversity. This approach has been successfully used in a variety of works \cite{stylegannada,diffusionclip,vqganclip}

\textbf{Classifier-Free Guidance}, \cite{cfg} observes that by using Bayes rule we can eliminate the necessity of an external classifier. By training the same model $\Prob_\theta$ to support both conditional and unconditional generation with conditioning dropout, we can thus rewrite the second term in Equation \ref{eq:classifier-guidance} as $\Prob_\theta(c|x) \propto \frac{\Prob_\theta(x|c)}{\Prob_\theta(x)}$. Then, the sampling is performed according to the probability: 

\begin{equation}\label{eq:cfg_sampling}
\widehat{\Prob_\theta}(x|c) \propto \frac{\Prob_\theta(x|c)^{\gamma}}{\Prob_\theta(x)^{\gamma-1}}. 
\end{equation} 

Modeling the diffusion process with $\widehat \Prob_\theta(x|c)$ effectively means predicting the PDF of the sample noise $\epsilon_t$ as

\begin{equation}
\label{eq:diffusion-cfg}
    \log \widehat{\Prob_\theta}(\epsilon_t|x_{t+1},c) = \gamma \log \Prob_\theta(\epsilon_t | x_{t+1}, c) - (\gamma - 1) \log \Prob_\theta(\epsilon_t | x_{t+1}).
\end{equation}

An important tool with diffusion models is \textbf{Negative Prompting} \cite{compositionalprompts,negprompt,crowson2022vqgan,stablediffusion}. We can rewrite Equation \ref{eq:diffusion-cfg} as

\begin{equation}
    \log \widehat{\Prob_\theta}(\epsilon_t|x_{t+1},c) = \log \Prob_\theta(\epsilon_t | x_{t+1}) + \gamma \big(\log \Prob_\theta(\epsilon_t | x_{t+1}, c) - \log \Prob_\theta(\epsilon_t | x_{t+1})\big)
\end{equation}

Aside from its probabilistic interpretation, this equation also represents a vector arithmetic operation in latent space: we take a step of size $\gamma$ away from the unconditional vector in the direction of the conditioning. Semantic vector linear arithmetic has proven to be effective in many situations in vision: striking examples have been generated by interpolations in GANs or diffusion models \cite{editgan,ddim,neuralphotoshop}. 

Moreover, the initial point does not have to be the unconditional latent, but any representation we want to move away from. We can introduce the "negative conditioning" or "negative prompt" $\overline{c}$, as well as a generalized equation resulting in Equation \ref{eq:diffusion-cfg} when $\overline{c}=\varnothing$:

\begin{equation}
    \log \widehat{\Prob_\theta}(\epsilon_t|x_{t+1}, c, \overline{c}) = \log \Prob_\theta(\epsilon_t | x_{t+1}, \overline{c}) + \gamma \big(\log \Prob_\theta(\epsilon_t | x_{t+1}, c) - \log \Prob_\theta(\epsilon_t | x_{t+1}, \overline{c})\big)
    \label{eq:negative-prompting}
\end{equation}

\subsection{Classifier-Free Guidance of Language Models}
\label{sct:methodology:guidance-llm}

To apply Classifier-Free Guidance to language models, we first have to define the semantic space to operate in. As demonstrated in \cite{word2vec,glove} and \cite{bert,gpt1}, word embeddings and sentence embeddings have strong semantic structures. This makes the logits of token predictions a good choice of our latent space, due to its linear relationship with the last hidden layer. Using the logits avoids network editing \cite{leace} and is architecture agnostic.

Next, we need to define what is considered conditioning, $c$, in decoder-only language models. In the common situations, a user provides a \textit{prompt} $c$ which can be a context, an instruction, or the beginning of some text, and uses a language model to sample a sequence of continuation tokens $w_i$ for the prompt $c$. Since a good continuation is expected to highly correlate to the prompt, we consider the prompt as our conditioning.

Similarly to Classifier Guidance \cite{dathathri2019plug, yang2021fudge, spangher2023sequentially}, we wish to generate a text $w$ which has a high likelihood of starting with $c$. We define the $\gamma$-reweighted distribution $\widehat{\Prob}(w|c) \propto \Prob(w)\cdot\Prob(c|w)^\gamma$, and approximate it with CFG as $\widehat{\Prob}(w|c) \propto \frac{\Prob(w|c)^\gamma}{\Prob(w)^{\gamma-1}}$

In the case of autoregressive language models modeling $\Prob_\theta(w)=\prod_i^N\Prob_\theta(w_i|w_{j<i})$, we can unroll the formulation and obtain Equation \ref{eq:cfg_sampling} again:

\begin{equation}
    \widehat{\Prob_\theta}(w | c) \propto \prod^T_{i=1} \widehat{\Prob_\theta}(w_i| w_{j<i}, c) \propto \prod^T_{i=1}\frac{\Prob_\theta(w_i|w_{j<i},c)^\gamma}{\Prob_\theta(w_i| w_{j<i})^{\gamma-1}} \propto \frac{\Prob_\theta(w|c)^\gamma}{\Prob_\theta(w)^{\gamma-1}}
    \label{eq:lm-cfg}
\end{equation}

While conditioned diffusion models cannot predict unconditioned distributions without extra training, language models handle both $\Prob_\theta(w|c)$ and $\Prob_\theta(w)$ naturally due to being trained on finite context windows. Being able to drop the prefix $c$ is a natural feature. 
We can thus sample the next $i$-th token $w_i$ in the logits space:

\begin{equation}
    \log \widehat{\Prob_\theta}(w_i|w_{j<i},c) = \log \Prob_\theta(w_i | w_{j<i}) + \gamma \big(\log \Prob_\theta(w_i | w_{i<j}, c) - \log \Prob_\theta(w_i | w_{j<i})\big)
\end{equation}


This formulation can be extended to accomodate ``negative prompting'', as in Equation \ref{eq:negative-prompting}. Negative prompting as applied in autoregressive LMs will be further addressed in Section \ref{sct:experiments:negative-prompting}. Now, we will continue on to the next section, where we introduce our experiments. In this section, we will explore the effects of CFG on different variations of prompting.

\section{Experiments}
\label{sct:experiments}

In this section we show that Classifier-Free Guidance reliably boosts performance across a variety of common prompting approaches. In Section \ref{sct:experiments:benchmarks} we show that CFG boosts zero-shot performance on a variety of standard NLP benchmarks, including achieving state-of-the-art performance on LAMBADA with LLaMa-7B. In Section \ref{sct:experiments:cot} we apply CFG to  \textit{Chain-of-Thought prompts} \cite{nye2022show,COT} an approach to allows the model to reason first before answering the question. Next, we test the performance of CFG on \textit{text-to-text generation prompts} in Section \ref{sct:experiments:text-to-text-generation}. Finally, we show in Section \ref{sct:experiments:negative-prompting} that CFG can be applied to \textit{assistant} prompts (i.e. prompts with system-instructions).

\subsection{Basic Prompting: Zero-Shot Prompts}
\label{sct:experiments:benchmarks}

To test \textit{basic, zero-shot prompting}, we consider a suite of zero-shot benchmarks implemented in the Language Model Evaluation Harness \cite{eval-harness}, which includes close-book QA \cite{auer2023sciqa, joshi2017triviaqa}, common sense reasoning tasks \cite{zellers2019hellaswag, sakaguchi2021winogrande, clark2019boolq, bisk2020piqa, cobbe2021training, basu2020aqua, arc}, and sentence completion-tasks \cite{paperno2016lambada}. In these settings, the desired completions are short (often 1-2 tokens), so risks of meandering \cite{spangher2023sequentially} or degradation \cite{holtzman2019curious} are low. We hypothesize that the main impact of CFG in these settings will be to reduce variance in output choices, as we explore more in Section \ref{sct:explanation}.

We evaluate the GPT-2 model family\cite{radford2019language}, the Pythia model family \cite{biderman2023pythia} and the LLaMA model family\cite{llama} using different guidance strengths across a range of standard NLP benchmarks using EleutherAI's Language
Model Evaluation Harness \cite{eval-harness} and implement CFG by starting the unconditional prompt at the last token of the initial prompt. The results are shown in Table \ref{tab:benchmark}. For better visualization, the charts for the GPT2 models, the Pythia models and the LLaMA models over the standard benchmarks are also shown in Figure \ref{fig:gpt2-split}, \ref{fig:pythia-split}, and \ref{fig:llama-split}, respectively. We observe that except ARC (challenge) and Winogrande, the boost of performances from CFG is nontrivial and consistent. The reasons for these discrepancies are still unknown.

Furthermore, we note that even the smallest LLaMA 7B model achieves $81\%$ accuracy in Lambada (OpenAI) zero-shot benchmark with $\gamma = 1.5$, outperforming the current SOTA (zero-shot) of PaLM-540B ($77.9\%$). Despite the fact that CFG almost doubles the computation during inference, the comparison is still noteworthy given that other models with comparable performances on Lambada (OpenAI) have much more parameters and would still require more compute than LLaMA 7B with CFG. Taken together, we show that CFG increases performance in basic prompting settings significantly. 

\begin{figure}[t]
    \centering
    \subfloat[][]{
    \begin{tabular}{lllll}
        \toprule
        {} &        ARC-c &        ARC-e &        BoolQ &    HellaSwag \\
        \midrule
        GPT2-small &   22.7 / \textbf{ 23.0} &   39.5 / \textbf{ 42.1} &   48.7 / \textbf{ 57.0} &   31.1 / \textbf{ 31.9}\\
        GPT2-medium &  \textbf{ 25.0} /  23.9 &   43.6 / \textbf{ 47.6} &   58.6 / \textbf{ 60.1} &   39.4 / \textbf{ 40.9}\\
        GPT2-large &  \textbf{ 25.1} /  24.7 &   46.6 / \textbf{ 51.0} &   60.5 / \textbf{ 62.1} &   45.3 / \textbf{ 47.1}\\
        GPT2-xl &   28.5 / \textbf{ 30.0} &   51.1 / \textbf{ 56.5} &   61.8 / \textbf{ 62.6} &   50.9 / \textbf{ 52.4}\\
        \midrule
        Pythia-160M &  \textbf{ 23.5} /  23.0 &   39.5 / \textbf{ 42.2} &   55.0 / \textbf{ 58.3} &   30.1 / \textbf{ 31.2}\\
        Pythia-410M &  \textbf{ 24.1} /  23.8 &   45.7 / \textbf{ 50.3} &   60.6 / \textbf{ 61.2} &   40.6 / \textbf{ 41.6}\\
        Pythia-1B &   27.0 / \textbf{ 28.0} &   49.0 / \textbf{ 54.9} &   60.7 / \textbf{ 61.8} &   47.1 / \textbf{ 48.9}\\
        Pythia-1.4B &   28.6 / \textbf{ 29.6} &   53.8 / \textbf{ 59.6} &   63.0 / \textbf{ 63.8} &   52.1 / \textbf{ 54.3}\\
        Pythia-2.8B &   33.1 / \textbf{ 34.5} &   58.8 / \textbf{ 65.4} &   64.7 / \textbf{ 64.7} &   59.3 / \textbf{ 61.9}\\
        Pythia-6.9B &   35.2 / \textbf{ 36.1} &   61.3 / \textbf{ 67.4} &   63.7 / \textbf{ 64.6} &   64.0 / \textbf{ 66.5}\\
        Pythia-12B &   36.9 / \textbf{ 38.7} &   64.1 / \textbf{ 72.6} &   67.6 / \textbf{ 67.8} &   67.3 / \textbf{ 69.6}\\
        \midrule
        LLaMA-7B &   41.5 / \textbf{ 43.9} &   52.5 / \textbf{ 58.9} &  \textbf{ 73.1} /  71.8 &   73.0 / \textbf{ 76.9}\\
        LLaMA-13B &   47.8 / \textbf{ 54.2} &   74.8 / \textbf{ 79.1} &  \textbf{ 78.0} /  75.8 &   79.1 / \textbf{ 82.1}\\
        LLaMA-30B &   52.9 / \textbf{ 57.4} &   78.9 / \textbf{ 83.2} &  \textbf{ 82.7} /  80.0 &   82.6 / \textbf{ 85.3}\\
        LLaMA-65B &   55.6 / \textbf{ 59.0} &   79.7 / \textbf{ 84.2} &  \textbf{ 84.8} /  83.0 &   84.1 / \textbf{ 86.3}\\

        \bottomrule
    \end{tabular}
    } \\ 
    
    \subfloat[][]{
    \begin{tabular}{llllll}
        \toprule
        {} &         PIQA &         SCIQ &   TriviaQA & WinoGrande &      Lambada \\
        \midrule
        GPT2-small &   62.5 / \textbf{ 63.8} &   64.4 / \textbf{ 70.8} &    5.5 / \textbf{  6.5} &  \textbf{ 51.6} /  50.5 &   32.6 / \textbf{ 44.6}\\
        GPT2-medium &   66.4 / \textbf{ 66.9} &   67.2 / \textbf{ 76.7} &    8.3 / \textbf{  9.3} &  \textbf{ 53.1} /  52.1 &   43.0 / \textbf{ 55.8}\\
        GPT2-large &   69.2 / \textbf{ 70.2} &   69.4 / \textbf{ 78.8} &   11.1 / \textbf{ 12.0} &  \textbf{ 55.4} /  54.4 &   47.7 / \textbf{ 60.5}\\
        GPT2-xl &   70.5 / \textbf{ 71.3} &   76.1 / \textbf{ 82.4} &   14.7 / \textbf{ 15.2} &  \textbf{ 58.3} /  55.6 &   51.2 / \textbf{ 62.5}\\
        \midrule
        Pythia-160M &   61.4 / \textbf{ 62.1} &   67.0 / \textbf{ 75.4} &    4.1 / \textbf{  5.3} &  \textbf{ 52.3} /  51.1 &   32.8 / \textbf{ 47.4}\\
        Pythia-410M &   67.1 / \textbf{ 67.8} &   72.1 / \textbf{ 79.0} &    7.9 / \textbf{  9.1} &  \textbf{ 52.9} /  50.7 &   51.3 / \textbf{ 64.0}\\
        Pythia-1B &   69.2 / \textbf{ 70.5} &   76.0 / \textbf{ 82.9} &  \textbf{ 12.3} /  12.3 &  \textbf{ 53.9} /  51.5 &   56.2 / \textbf{ 69.0}\\
        Pythia-1.4B &   71.1 / \textbf{ 72.5} &   79.4 / \textbf{ 85.1} &   15.9 / \textbf{ 15.9} &  \textbf{ 57.4} /  56.0 &   61.6 / \textbf{ 72.7}\\
        Pythia-2.8B &   73.6 / \textbf{ 75.8} &   83.3 / \textbf{ 88.2} &  \textbf{ 22.1} /  20.9 &  \textbf{ 60.1} /  57.9 &   64.6 / \textbf{ 76.5}\\
        Pythia-6.9B &   76.3 / \textbf{ 77.4} &   84.3 / \textbf{ 89.7} &  \textbf{ 28.2} /  27.2 &  \textbf{ 61.1} /  60.3 &   67.1 / \textbf{ 78.8}\\
        Pythia-12B &   77.0 / \textbf{ 78.4} &   87.7 / \textbf{ 91.9} &  \textbf{ 33.4} /  32.1 &  \textbf{ 65.0} /  63.4 &   70.4 / \textbf{ 80.6}\\
        \midrule
        LLaMA-7B &   77.4 / \textbf{ 79.8} &   66.3 / \textbf{ 75.4} &  \textbf{ 56.0} /  52.7 &  \textbf{ 67.1} /  65.5 &   73.6 / \textbf{ 81.3}\\
        LLaMA-13B &   80.1 / \textbf{ 80.9} &   91.1 / \textbf{ 95.1} &  \textbf{ 62.4} /  59.8 &  \textbf{ 72.8} /  71.5 &   76.2 / \textbf{ 82.2}\\
        LLaMA-30B &  \textbf{ 82.3} /  82.3 &   94.3 / \textbf{ 96.4} &  \textbf{ 69.7} /  67.9 &  \textbf{ 75.8} /  74.1 &   77.5 / \textbf{ 83.9}\\
        LLaMA-65B &   82.3 / \textbf{ 82.6} &   95.1 / \textbf{ 96.6} &  \textbf{ 73.3} /  71.8 &  \textbf{ 77.4} /  76.1 &   79.1 / \textbf{ 84.0}\\
        \bottomrule
    \end{tabular}
    }
    \caption{Results of general natural language benchmarks. In each cell, the first value is the result for $\gamma = 1$ (baseline) and the second value is the result for $\gamma = 1.5$ (ours). LLaMA 7B with CFG on Lambada zero-shot already outperforms vanilla PaLM 540B, Chinchilla 70B, and GPT-3 175B, tops the \href{https://paperswithcode.com/sota/language-modelling-on-lambada}{SOTA leaderboard} for Lambada zero-shot as of June 26th, 2023}
    \label{tab:benchmark}
\end{figure}

\subsection{Deliberative Prompting: Chain-of-Thought}
\label{sct:experiments:cot}

A variation on \textit{basic prompting} has emerged recently called \textit{Chain-of-Thought (CoT) prompting} \cite{COT}. In this setting, the model is prompted to generate a series of reasoning steps before giving an answer to the task: i.e. $p(w_{cot}, w_a| w_p)$, where $w_{cot} = w_{p+1}...w_{c-1}$ and $w_a$ is the answer. $w_{cot}$ is designed to mimic the human reasoning or deliberation process. CoT has been shown to perform well in complex reasoning tasks that can not be fully addressed by model- or data-scaling ~\cite{rae}, however, as observed by~\cite{COT}, long reasoning chains can diverge and either do not generate correct answers, or do not follow the expected result structure given by the prompt.

This setting poses a variation on the prior \textit{base-case} setting: now, the continuation $w_c = \left[w_{cot}, w_{a}\right]$ is expected to be longer than 1-2 tokens. We hypothesize that compared to basic zero-shot prompting explored in Section \ref{sct:experiments:benchmarks}, CFG will \textit{also} be able to enforce better reasoning chains with less drift.

\begin{figure}[!ht]
    \centering
    \includegraphics[width=.4\linewidth]{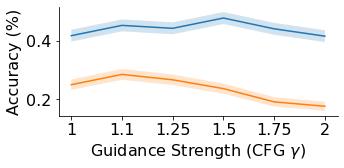}
    \includegraphics[width=.4\linewidth]{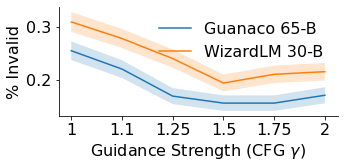}
    \caption{CFG impact on chain-of-thought prompting with respect to GSM8K dataset. For small CFG values, using CFG increases the percentage of chains which end in a valid answer structure while increasing the model accuracy. For large values the invalid percentage remains small but the accuracy drop.  }\label{fig:cot_GSM8K}
\end{figure}

 We evaluate the effectiveness of our proposed CFG method with respect to chain-of-thought prompting on two arithmetic reasoning tasks: GSM8K~\cite{gsm8k} and AQuA~\cite{aqua}. We follow~\cite{self-consistency} few-shot prompt and parsing setting, with respect to two open source LLM models: WizardLM-30B~\cite{xu2023wizardlm} and Guanaco-65B~\cite{guancano}.
As can be seen in Figure~\ref{fig:cot_GSM8K},~\ref{fig:cot_aqua}, using CFG increases the percentage of CoT which results in a valid answer that could be parsed. For low guidance strengths, this results in boosting the model performances. However, for large values, although the model returns more valid results, the quality of the chains is also impacted, and overall the model performances degrade. A qualitative comparison is provided in Table~\ref{tbl:cot_qual1},~\ref{tbl:cot_qual2}.  

We have only scratched the surface of exploring CFG's interactions with CoT; for instance, instead of upweighting just $w_p$, we might upweight $w_p, w_{cot}$, or other variations. We anticipate in future work being able to more fully test variations of CFG-weighting on different parts of the CoT process.

\subsection{Text-to-Text Prompts: Generation}
\label{sct:experiments:text-to-text-generation}
In contrast to \textit{basic prompting} and \textit{CoT-prompting}, where we ultimately expect a short answer, $w_a$, many settings require lengthier continuations. In this section, we study a prompt setting where the quality of answers are highly dependent the ability to stay on target over long sequences of text (both prompt, $w_p$ and continuation, $w_c$). Here we focus on code generation, and in Appendix \ref{app:machine-translation} we report results on machine translation. We hypothesize that, in contrast to Sections \ref{sct:experiments:benchmarks} and \ref{sct:experiments:cot}, these tasks require longer-form completions, which Classifier-Free Guidance's effectiveness in enforcing adherences to many different parts of the prompt. 

\subsubsection{Program synthesis evaluations}\label{sct:experiments:text-to-text-generation:program_synthesis}
Computer programs represent an important language-modeling case, as formal language differs from natural language in many ways including the use of well-defined structures. Testing Classifier-Free Guidance on code-related tasks improves the robustness of our hypothesis over different distributions of data. In the exploratory experiments, we prompt GPT-J \cite{gpt-j} and CodeGen-350M-mono \cite{nijkamp2023codegen} for small-scale code generations and observe positive results results (see Appendix \ref{appendix:additional-exp}). And then we perform a thorough evaluation on the HumanEval benchmark \cite{chen2021codex}.

\subsubsection{HumanEval benchmark}

To systematically investigate the impact of Classifier-Free Guidance on code completion abilities, we evaluate models using different CFG strengths on HumanEval benchmark \cite{chen2021codex}. HumanEval benchmark contains $164$ coding tasks in Python where the prompts are given by a function signature and a docstring. The model will generate continuations of the prompt, and the resulting programs will be tested against a set of unit tests for each task which evaluate the correctness of Python programs. We choose CodeGen-350M-mono, CodeGen-2B-mono and CodeGen-6B-mono (\cite{nijkamp2023codegen}) which are specialized in Python program synthesis.\footnote{\textit{Note: CodeGen-16B-mono is omitted due to the compute constraint.}}

Various CFG strengths \footnote{$\gamma = 1.0, 1.1, 1.25, 1.5, 1.75, 2.0$} are tested on $3$ different temperatures $0.2, 0.6, 0.8$ with the evaluation metrics being pass@$k$ for $k=1, 10, 100$ \footnote{The definition of pass@$k$ according to \cite{chen2021codex}: ``$k$ code samples are generated
per problem, a problem is considered solved if any sample passes the unit tests, and the total fraction of problems
solved is reported."}. Here we show the results for temperature$=0.2$ in Table \ref{tab:codegen}. The full results are summarized in Appendix \ref{appendix:charts:humaneval} in Table \ref{tab:codegen-350}, \ref{tab:codegen-2b} and \ref{tab:codegen-6b} and Figure \ref{fig:codegen350}, \ref{fig:codegen2b} and \ref{fig:codegen6b}.

\begin{table}[t]
    \centering
    \begin{tabular}{llllllllll}
        \toprule
         & \multicolumn{3}{c}{CodeGen-350M} & \multicolumn{3}{c}{CodeGen-2B} & \multicolumn{3}{c}{CodeGen-6B} \\
         \cmidrule(r){2-4} \cmidrule(r){5-7} \cmidrule(r){8-10}
        $\gamma$ & k=1 & k=10 & k=100 & k=1 & k=10 & k=100 & k=1 & k=10 & k=100 \\
        \midrule
        1.0 & 11.0\% & 17.0\% & \textbf{22.0\%} & 19.5\% & 25.5\% & \textbf{29.8\%} & 19.5\% & 25.5\% & \textbf{29.8\%} \\
        1.1 & \textbf{11.8\%} & \textbf{18.1\%} & 20.1\% & 20.4\% & 25.4\% & 28.0 & 20.4\% & 25.4\% & 28.0\% \\
        1.25 & 11.4\% & 17.3\% & 18.9\% & 19.7\% & 25.4\% & 28.0 & 19.7\% & 25.4\% & 28.0\% \\
        1.5 & 10.9\% & 16.7\% & 18.3\% & \textbf{20.9\%} & \textbf{26.7\%} & 29.2\% & \textbf{20.9\%} & \textbf{26.7\%} & 29.2\\
        1.75 & 10.3\% & 16.0\% & 18.2\% & 20.4\% & 26.2\% & 28.6\% & 20.4\% & 26.2\% & 28.6\% \\
        2.0 & 8.6\% & 14.6\% & 17.6\% & 16.5\% & 22.4\% & 24.4\% & 16.5\% & 22.4\% & 24.4\% \\
        \bottomrule\\
    \end{tabular}
    \caption{CodeGen results with temperature$=0.2$. CFG in nearly all cases increases performance, but the optimal $\gamma$ value varies.}
    \label{tab:codegen}
\end{table}

We observe that low CFG ($\gamma \leq 1.5 $) increases the pass@$1$ rate uniformly\footnote{Note that the effect of low CFG on the pass@$1$ rate is consistent with the results of the general benchmarks in the previous section.}. High CFG ($\gamma \geq 1.5$) leads to a deterioration of performance. We also note that the improvement from CFG diminishes or harms performance at pass@$k$ at high $k$.

\begin{figure}[htp]
  \centering
    \begin{minipage}{.45\linewidth}
        \centering
        \includegraphics[width=\linewidth]{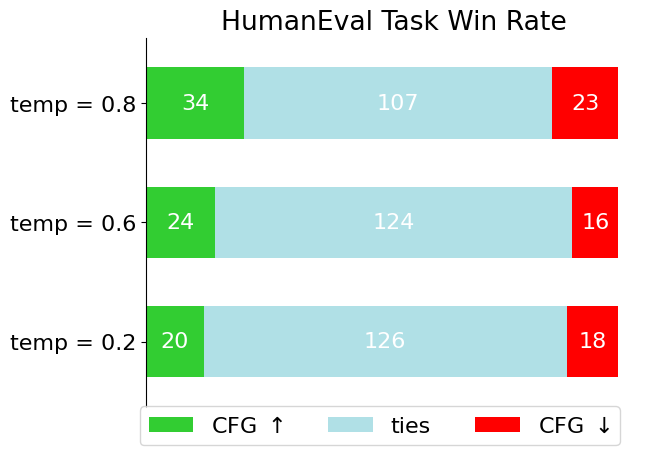}
        \caption{HumanEval task count comparison between $\gamma = 1, 1.25$ for CodeGen-350M-mono}
        \label{tab:codegen-350-task-count}
    \end{minipage} \hfill
    \begin{minipage}{.45\linewidth}
        \centering
        \includegraphics[width=\linewidth]{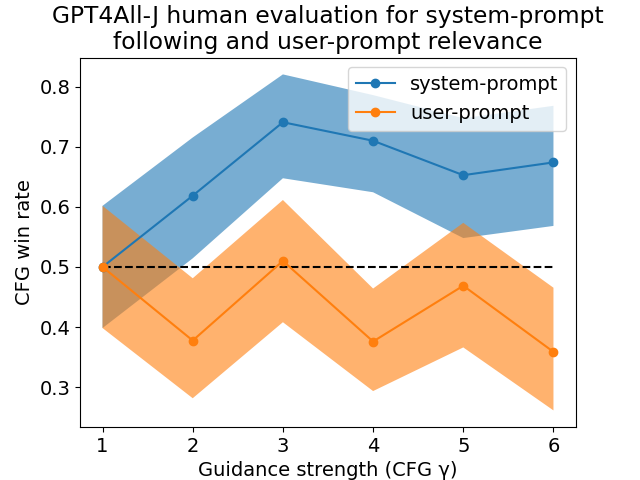}
        \caption{Evaluators (611 votes, 71 unique voters) significantly preferred the system-prompt with CFG (max at $\gamma=3$) . The user-prompt relevance, not subject to CFG, did not degrade until $\gamma\ge 4$, showing a clear win without tradeoff at $\gamma=3$.}
        \label{fig:gpt4all-human}
    \end{minipage}
\end{figure}

To further investigate the effect of CFG, we break down the pass@$1$ evaluations on CodeGen-350M-mono for $\gamma = 1, 1.25$ task-by-task \footnote{See the scatter plot at temperature $0.2, 0.6, 0.8$ in appendix, Figure \ref{fig:humaneval-tasks-0.8}, \ref{fig:humaneval-tasks-0.6}, \ref{fig:humaneval-tasks-0.2}}. We notice that the number of tasks where CFG outperforms is still more than the one where CFG underperforms for all temperatures $0.2, 0.6, 0.8$ (See Table \ref{tab:codegen-350-task-count}).

We also find that without CFG, many tasks exhibit small nonzero passing rates while having $0\%$ rate with CFG. This explains the decreasing improvement of CFG in pass@$k$ for large $k$, as larger $k$ significantly boosts the passing rate of difficult tasks where the rates are low but nonzero.

Overall, the consistent improvement on pass@$1$ rates and the reduced effect on pass@$100$ rates support our hypothesis that CFG strengthens the adherence to the prompt at the small cost of reduced variability and creativity.

\subsection{Negative Prompting: Improving Assistants}
\label{sct:experiments:negative-prompting}

Finally, we explore an addition to Classifier-Free Guidance called \textit{negative prompting}. With negative prompting, the user specifies what they do \textit{not} want in the output (e.g. ``low resolution'', ``bad hands, bad anatomy, amateur drawing'' in text-to-image), which is then used to improve generation quality.

We explore this idea in the context of chatbots. Chatbots give us a setting where the \textit{prompt} is expanded into a \textit{multi-stage prompt}\footnote{We note that this extension to \textit{basic-prompting} stands as a mirror to \textit{CoT-prompting}'s extension (Section \ref{sct:experiments:cot}). In \textit{CoT-prompting}, the \textit{continuation} is expanded to a \textit{multi-stage completion}; here, the \textit{prompt} is expanded.}. In chatbots, the language model is prompted with a two-part prompt: (1) the instruction, $w_s$ (sometimes called "system prompt") which may give contextual information (e.g. the ``current date''), or behavioral guidelines (e.g. style, alignment, persona, etc.); and (2) $w_p$, the user-prompt, or the user's query. See Table \ref{tbl:example} for an example. Adherence becomes an even greater challenge, as our initial explorations shows. We observe systems like Alpaca \cite{alpaca,vicuna,gpt4all} often ignore changes to their default system-prompt, and may even expose models to attacks like prompt injection \cite{greshake2023more}. 

We explore CFG with negative prompting to increase the success rate of different system prompts. We set the negative prompt $\overline{c}$ to be the default system-prompt for the models we use (i.e. ``The prompt below is a question to answer, a task to complete, or a conversation to respond to; decide which and write an appropriate response.'') and set $c$ to be the edited prompt (e.g. ``The prompt below is a question to answer, a task to complete, or a conversation to respond to; decide which and write \textit{a sad} response.''). This approach not only makes the sampling more prompt-aware in general, but directly emphasizes the difference between \textit{our} system-prompt and the model's default system-prompt. 

To test this approach with chatbots, we generate system-prompts, $n_{c} = 25$, and user-prompts, $n_{p} = 46$, and sample $1740$ random combinations of them. An example is shown in Table \ref{tbl:example} (in Appendix \ref{sec:prompts} we include the full list of $c$ and $p$ we use). We use \texttt{GPT4All-J v1.3-jazzy} to generate two completions for each sampled combination: the first is sampled without CFG, and the second is sampled with CFG, with a guidance strength randomly chosen $\in$ {1,2,3,4,5,6}. Our hypothesis is that CFG increases system-prompt following, ideally without hurting the relevance to the user input. 

We run a human preference study on our sampled continuations, where participants are shown both, blindly, and asked to assess two things: A.  which output better follows the system-prompt, $c$ and B. which output better follows the user-prompt $p$. Our results in Figure \ref{fig:gpt4all-human} shows compelling evidence that CFG emphasized the difference between $c$ and $\overline{c}$ more than sampling with $c$ alone. There is a clear peak at $\gamma=3$ with 75\% of system-prompt following preference over $\gamma=1$ and undegraded user-prompt relevance (52\%).

\section{Computational Cost Analysis}\label{sct:cost-analysis}

In the previous section we showed improvements across a wide array of benchmarks and contexts. However, since classifier-free guidance requires two passes through the network, users who are compute-constrained rather than VRAM constrained might wonder if CFG is interesting to them at all, and if they should not run a model twice as big instead.

To answer this question, we calculate the FLOP for each of the benchmark experiments that we ran in Section \ref{sct:experiments:benchmarks}. We then compare across model sizes, with and without CFG. We conclude with the surprising finding that, across 5 out of 9 tasks, there there is a statistically \textit{insignificant difference} between using CFG and using vanilla prompting with a model of twice the size at $p=.01$, according to ANCOVA regression analysis \cite{rutherford2011anova}. Of the significantly different tasks, 2 favor CFG and 2 favor vanilla. See Appendix \ref{app:flops}, specifically Figure \ref{fig:flops}, for more details. 

In other words, and most significantly, this indicates that, overall, a model using CFG can generally perform just as well as a model twice as large. This has enormous implications for training budgets and inference latency due to limited VRAM usage, which we seek to explore in future work.

\section{Explaining the Success of Classifier-Free Guidance}
\label{sct:explanation}

In this section, we try to derive insights on the impact of Classifier-Free Guidance on generation, both quantitatively and qualitatively. We sample a dataset of $32,902$ datapoints from the P3 dataset \cite{sanhmultitask} and use the Falcon-7b-Base model family \cite{falcon40b} as an exploratory model. Our goal is to analyze the logit distributions -- we describe how in the following sections. Many of our comparisons are done with reference to an instruction-tuned model, for which we use the Falcon-7b-Instruct version. We replicate our findings on other models and datasets as well: the Open-Assistant Dataset \cite{kopf2023openassistant} and Redpajama-3b model family\footnote{\url{https://www.together.xyz/blog/redpajama}}.

\begin{figure}
    \centering
    \subfloat[Entropy of logits for the vanilla prompted distribution $\Prob(y|x)$, the unprompted distribution, $\Prob(x)$, the CFG-$\gamma=1.5$ distribution and an instruction-tuned model $\Prob_\text{instruct}(y|x)$.]{\includegraphics[width=.45\linewidth]{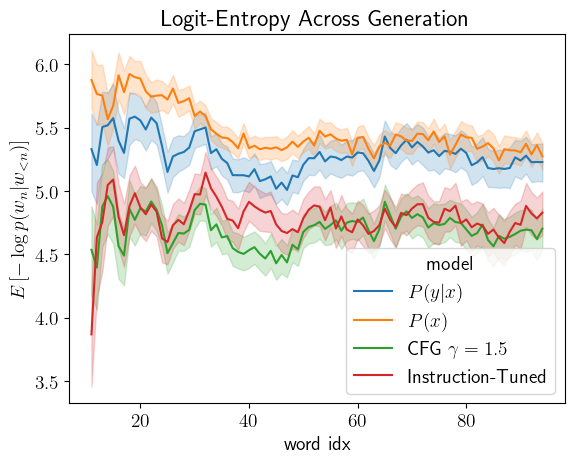}
    \label{fig:entropy}
    }
    \hfill
    \subfloat[Number of tokens overlapping in top-p=90\% of vocabulary distributions between that of: CFG, that of the vanilla prompted model, $p(y|x)$, and that of the unprompted model, $\Prob(x)$.]{\includegraphics[width=.45\linewidth]{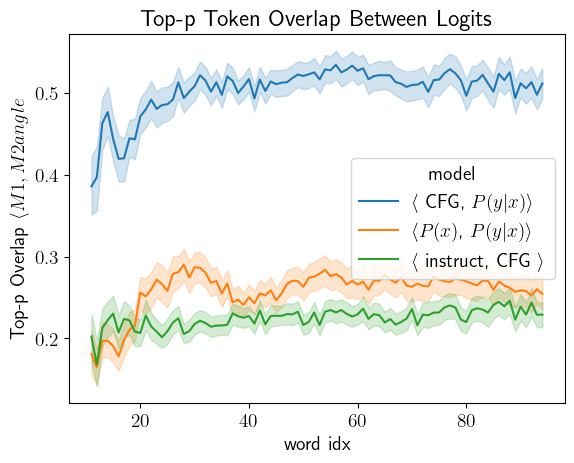}
    \label{fig:topp}}
    \caption{We show into how CFG alters the logit distribution of the vanilla prompted model, $\Prob(y|x)$. CFG lowers the entropy to a level roughly similar to instruction-tuned model variant. CFG shares roughly 50\% of the tokens in top-p=$0.9$ as the vanilla $\Prob(y|x)$ model.}
    \label{fig:single-model-comparisons}
\end{figure}

\subsection{Classifier-Free Guidance's Effect on Sampling Entropy}
\label{sct:explanation:entropy}

We suspect that CFG, by focusing $\Prob(y|x)$ on the prompt, will reduce the entropy of the logit distribution. CFG entropy distribution is significantly lower across generation time-steps  vanilla prompting, with a mean of 4.7 vs. 5.4. (See Figure \ref{fig:entropy}).
The effect this has is to restrict the number of tokens in the top-p=90\% of the vocabulary distribution (See in Figure \ref{fig:topp}). We do observe qualitatively, shown in Section \ref{sct:explanation:visualizing}, that the top tokens to not shift too much, but they do re-order to some extent, which shows that CFG is not simply having the same effect as the temperature parameter.

\subsection{CFG's Relation to Instruction Tuning}
\label{sct:explanation:instruction}

Our next question: \textit{how} is Classifier-Free Guidance affecting the vocabulary distribution? We attempt to answer this question quantitatively, hypothesizing that CFG has similar effects to instruction-tuning, which we assume trains a model to focus on the prompt.  We find that both CFG and Instruction-Tuned model variants have similar entropies across generation samples. However, as shown in Figure \ref{fig:topp} the vocabulary distributions across our samples are largely not overlapping. 

\begin{figure}
\centering
\subfloat[Correlation between the perplexities of each model on P3.]{
\begin{tabular}{llll}
\toprule
{} & PPL $p(y|x)$ & PPL cfg & PPL instruct \\
\midrule
PPL $p(y|x)$ &  \cellcolor{blue!100} \textcolor{white}{1.0} &  \cellcolor{blue!87} \textcolor{white}{0.94} &  \cellcolor{blue!65} \textcolor{white}{0.83} \\
PPL cfg &  \cellcolor{blue!87} \textcolor{white}{0.94} &  \cellcolor{blue!100} \textcolor{white}{1.0} &   \cellcolor{blue!39} \textcolor{white}{0.7} \\
PPL instruct  &  \cellcolor{blue!65} \textcolor{white}{0.83} &   \cellcolor{blue!39} \textcolor{white}{0.7} &  \cellcolor{blue!100} \textcolor{white}{1.0} \\
\bottomrule
\end{tabular}
}\hspace{2em}
\subfloat[Correlation between the perplexity and similarity between Instruction-Tuned and CFG.]{
\begin{tabular}{lll}
\toprule
{} &        $r_s$ (sim) &     $p$-val. \\
\midrule
PPL $p(y|x)$          &   \cellcolor{blue!10}0.01 &   0.2 \\
PPL cfg               &  \cellcolor{red!40}-0.04 &  <.001 \\
PPL instruct  &   \cellcolor{blue!40}0.04 &  <.001 \\
\bottomrule
\end{tabular}
}
\caption{We seek to identify \textit{when} CFG is similar to instruction-tuning. Models mostly agree on the difficulty of input sentences, and in cases where they do not, CFG and Instruction-tuning have similar top-p overlaps.}
\label{tbl:ppls_of_diff_models}
\end{figure}

We find that, overall, our hypothesis about the similarity is wrong: CFG is not having a similar effect on the vocabulary logits as instruction-tuning. To explore, we seek to derive insight from edge-cases where it does. We look for characteristics to explain when CFG \textit{is} similar to Instruction-Tuning (in terms of top-p overlap). One case pops out: when the prompt is longer, CFG agrees more -- we observe a significant spearman correlation of $r_s=.05$ between prompt-length and Instruction/CFG agreement. We also observe small but significant correlations between perplexity and agreement. As shown in Table \ref{tbl:ppls_of_diff_models}, harder phrases for Instruction-Tuned models are typically where CFG and Instruction-Tuned models align. We conclude that CFG is altering the model in ways that might complement instruction-tuning, opening the door to future explorations.

\subsection{Visualizing Classifier-Free Guidance}
\label{sct:explanation:visualizing}

Finally, we provide qualitative insights into the reordering of the vocabulary, after Classifier-Free Guidance is applied. We note that the Equation can be rewritten as 

\begin{equation}
    \log \Prob_\gamma(w_t|w_{<t},c) = \log \Prob(w_t| w_{<t},\overline{c}) + \gamma (\log \Prob(w_t| w_{<t},c) - \log \Prob(w_T|w_{<t},\overline{c})
\label{eq:cfg-2}
\end{equation}

We propose, at each timestep, to visualize the vocabulary ranked by the difference $\log \Prob(w_t| w_{<t}) - \log \Prob(w_T| \hat{w})$. This shows the impact of the method, qualitatively, by revealing the tokens that are encouraged or discouraged the most. In Figure \ref{fig:viz}, we prompt a model with $c=$``The dragon flew over Paris, France'',$\overline{c}=\emptyset$ and observe that tokens about dragons and Paris get upweighted while tokens about other locations (``Queensland''), dates (``1913''), or topics (``hostages'', ```voyages'') get downweighted. This confirms our initial assumptions, as we observe CFG encouraging tokens related to and discourages tokens unrelated to the prompt.

\begin{table}
  \centering
  \resizebox{\textwidth}{!}{%
  \begin{tabular}{llllllllllll}
    \toprule
    \textbf{current} & top1 & top2 & top3 & top4 & top5 & ... & bottom5 & bottom4 & bottom3 & bottom2 & bottom1 \\
    \midrule
    \textbf{\ France} & \ flipping & \textcolor{red}{\ destroying} & \ waking & \ stopping & \ causing & ... & \ guiName & ufact & \ Outs & kees & "\}]," \\
    \textbf{,} & \ \textcolor{red}{crashing} & \ \textcolor{red}{landing} & \ \textcolor{red}{soaring} & \ swoop & \ plummet & ... & soDeliveryDate & \ POLIT & \ Occupations & 568 & \textcolor{blue}{\ publishes} \\
    \textbf{\ landing} & \ neigh & \ invis & \ atop & \ overhead & \ \textcolor{red}{omin} & ... & \textcolor{blue}{\ quotas} & \textcolor{blue}{\ Russo} & \textcolor{blue}{\ Germans} & \textcolor{blue}{\ passports} & \textcolor{blue}{\ hostages} \\
    \textbf{\ on} & \ Buildings & \ skysc & \ rooft & \ Cheong & \ Plaza & ... & \ \begin{CJK}{UTF8}{gbsn}裏\end{CJK} & \ MFT & \begin{CJK}{UTF8}{gbsn}ゼ\end{CJK} & \ \begin{CJK}{UTF8}{gbsn}裏覚醒\end{CJK} & DragonMagazine \\
    \textbf{\ Notre} & \textcolor{OliveGreen}{\ Basil} & \ Mos & \textcolor{OliveGreen}{\ Cathedral} & \textcolor{OliveGreen}{\ Mosque} & \ Eugene & ... & \textcolor{blue}{\ voyage} & alach & urse & arb & sb \\
    \textbf{\ Dame} & \textcolor{OliveGreen}{\ Cathedral} & \ monument & \textcolor{OliveGreen}{\ cathedral} & \textcolor{OliveGreen}{\ Basil} & \ Mosque & ... & \textcolor{blue}{\ voyage} & aila & voy & aund & wk \\
    \textbf{\ Cathedral} & .," & ."[ & \ slowing & \ blocking & ortex & ... & \textcolor{blue}{\ ashore} & \ seaf & aund & \ Tact & \ Wanted \\
    \textbf{.} & \ \textcolor{red}{Dragon} & \textcolor{red}{\ dragons} & \ \textcolor{red}{dragon} & \textcolor{red}{Dragon} & \ \textcolor{red}{Dragons} & ... & \textcolor{blue}{\ 1915} & \textcolor{blue}{\ 1914} & \textcolor{blue}{\ 1944} & \textcolor{blue}{\ 1934} & \textcolor{blue}{\ 1913} \\
    \textbf{\ It} & \ swoop & \textcolor{red}{\ circled} & \ dart & \ hopped & \ bolted & ... & \ concludes & reads & \ reads & \ culmin & \ marks \\
    \textbf{\ circled} & \textcolor{red}{\ skysc} & \ pedestrians & \ architectural & \ hanging & \ skyline & ... & \textcolor{blue}{\ Newfoundland} & \textcolor{blue}{\ Ukrain} & \textcolor{blue}{\ Zamb} & \textcolor{blue}{\ Johnston} & \textcolor{blue}{\ Queensland} \\
    \textbf{\ Paris} & night & amura & rum & anim & animate & ... & \ prematurely & \ capit & \textcolor{blue}{\ bombed} & \ Mé & \ owing \\
    \textbf{\ a} & \ longer & \ while & \ long & \ awhile & \ length & ... & ims & chin & chel & ille & ller \\
    \textbf{\ bit} & \ longer & MORE & \ awhile & again & \ more & ... & \ prematurely & hof & nw & \ arri & trop \\
    \textbf{,} & \ startled & \textcolor{red}{\ feathers} & \textcolor{red}{\ dragon} & \textcolor{red}{\ wings} & \textcolor{red}{\ dragons} & ... & \ inval & \ \textcolor{blue}{Junction} & \ \textcolor{blue}{Palest} & endas & CVE \\
    \textbf{\ and} & \textcolor{red}{\ dragon} & \textcolor{red}{\ dragons} & \ golden & \textcolor{red}{\ Winged} & \ perched & ... & CVE & \ inval & \ \textcolor{blue}{Ukrain} & onet & \ \textcolor{blue}{Commodore} \\
    \textbf{\ then} & \textcolor{red}{\ dragon} & \ DRAG & \ \textcolor{red}{dragons} & \ neigh & \ \textcolor{red}{DRAGON} & ... & CVE & onet & \ Kear & TPS & Tags \\
    \textbf{\ flew} & ukong & \ skelet & rum & \ swoop & acles & ... & RG & thouse & NJ & \ 444 & \ programmes \\
    \textbf{\ over} & rium & \textcolor{blue}{\ Rockefeller} & \textcolor{blue}{\ Plaza} & \textcolor{blue}{\ Times} & \textcolor{blue}{\ Symphony} & ... & \ \textcolor{blue}{Brittany} & \ \textcolor{blue}{Newfoundland} & \textcolor{blue}{Balt} & \textcolor{blue}{isconsin} & \ \textcolor{blue}{Yugoslavia} \\
    \textbf{\ the} & \ Griffith & \ Zeus & \ Hag & \ Science & \ Raphael & ... & shire & \ \textcolor{blue}{Midlands} & \ \textcolor{blue}{frontier} & \ \textcolor{blue}{deserts} & \ Balkans \\
    \textbf{\ E} & BI & \ Rowe & \ ident & \ Methodist & allah & ... & \ \textcolor{blue}{coasts} & ento & bys & seys & \ Desire \\
    \textbf{iff} & \ Armory & \ Library & \ restrooms & \ Mansion & \ Mahmoud & ... & \ indo & onne & Off & itime & Norm \\
    \textbf{el} & \textcolor{OliveGreen}{\ restaurant} & \ Middle & \textcolor{OliveGreen}{\ restroom} & \textcolor{OliveGreen}{\ boutique} & \textcolor{OliveGreen}{\ museum} & ... & iband & \ \textcolor{blue}{throats} & \ centres & \ detach & rift \\
    \textbf{\ Tower} & \ Property & omin & \ Foundation & \ Creature & >" & ... & gee & \ thence & pheus & \ hither & \ favourable \\
    \textbf{.} & \textcolor{red}{\ dragons} & \textcolor{red}{\ dragon} & \textcolor{red}{\ Dragons} & \textcolor{red}{\ Dragon} & \textcolor{red}{\ DRAGON} & ... & \ \textcolor{blue}{1944} & \ \textcolor{blue}{1942} & \ Instrument & \textcolor{blue}{Balt} & \ \textcolor{blue}{1943} \\
    \textbf{\ Then} & \textcolor{red}{\ dragons} & \textcolor{red}{\ dragon} & \textcolor{red}{dragon} & \textcolor{red}{\ Dragons} & \textcolor{red}{Dragon} & ... & \ Manz & \ \textcolor{blue}{Hopkins} & CVE & \ \textcolor{blue}{Instrument} & \ \textcolor{blue}{Squadron} \\
    \textbf{\ it} & \textcolor{red}{\ dragon} & \textcolor{red}{\ dragons} & \ neigh & \ \textcolor{red}{Winged} & \ Draco & ... & CVE & udder & services & \ corrections & obbies \\
    \textbf{\ flew} & \ upro & ukong & rum & \ walked & \ …" & ... & INC & inary & lein & \ auxiliary & CVE \\
    \textbf{\ over} & \textcolor{blue}{\ Chinatown} & \textcolor{blue}{\ Financial} & Spider & \ tallest & \textcolor{blue}{Financial} & ... & \ warr &  & quickShip & \ \textcolor{blue}{Newfoundland} & \\
    \bottomrule\\
    \end{tabular}
    }
    \caption{Given the prompt \textbf{The dragon flew over Paris, France} we display, at each sampling step, the vocabulary ranked for $\Prob(w_t| w_{<t}) - \log \Prob(w_T| \hat{w})$ for the next step. We can see CFG encouraging tokens about \textcolor{red}{flying dragons} and \textcolor{OliveGreen}{Paris}, and discouraging other \textcolor{blue}{topics or regions}}
\label{fig:viz}
\end{table}

We find this visualization approach to be a useful prompt engineering tool, by using the new prompt under testing as $c$ and setting $\overline{c}$ as the current baseline prompt. The visualization shows the differential impact over the whole vocabulary on the next token prediction, in an interpretable way.

\section{Conclusion}

We have shown that Classifier-Free Guidance, which was originally conceived of in text-to-image applications, can be an effective way of increasing adherence to the prompt in autoregressive language modeling. In contrast to text-to-vision, CFG in autoregressive language modeling works out-of-the-box, without the need to further train the model. We have shown that CFG can boost performance across an array of canonical benchmarks in NLP that involve variations of the prompt: \textit{basic prompting}, \textit{chain-of-thought prompting}, \textit{text-to-text prompting} and \textit{chatbot prompting}. Finally, we sought to explain the effects of CFG by showing it decreased sampling entropy, but not in the same ways that Instruction-tuned models do. Ultimately, we leave for future work the exact effects that CFG is having, but we propose qualitative visualizations that confirm our intuitions around prompt adherence.

Our work also integrates into a growing body of inference techniques aimed at perturbing the logit distributions of an LM \cite{li2022contrastive, shi2023trusting}. We demonstrate that by doubling the inference FLOP using CFG brings performances of a model about twice the size. This allows training smaller models, which can be ran on smaller hardware, and are cheaper to train.

Our work faces the following limitations: CFG requires tweaking and exploration: $\gamma$ values that might work in one context (i.e. long-form generation) might be poorly suited for another context. It's also possible that CFG might be misused. We have not tested the effects of CFG if used in conjunction with malicious strategies for hacking language models, including but not limited to: prompt injection and prompts aimed at overriding alignment. It's possible that there are unforeseen effects induced by an increased adherence to parts of the prompt. We tried to explore this at length, both quantitatively and qualitatively, and we designed tasks that might reveal such behavior. However, we cannot conclude this method is risk-free. We advocate for standardized benchmarks aimed more squarely at language-model risk (including, possibly, pairs of models along with known prompt injections). Such standardized benchmarks could help us unit-test an advancement like CFG before releasing it into the wild. 

\subsection*{Acknowledgements}

We are grateful to Stability and CoreWeave for providing the compute to run the evaluations.

We also thank the volunteers who took part in the GPT4All experiment.

Alexander Spangher would like to thank Bloomberg News for a 4 year PhD fellowship that generously funds his research.

\bibliography{refs}

\newpage
\appendix
\addcontentsline{toc}{section}{Appendix} 
\part{Appendix} 
\parttoc 

\section{Author Contributions}

This work is a spontaneous collaboration between EleutherAI members and EleutherAI's Discord's members.

\paragraph{Guillaume Sanchez} came up with the initial theory, code and preliminary experiments, then reached EleutherAI in search for collaborators. He wrote the code for \ref{tab:benchmark} and associated figures, redacted Sections \ref{sct:methodology:guidance-t2i}, \ref{sct:methodology:guidance-llm}. He wrote the code and ran the GPT-J experiment mentioned in \ref{sct:experiments:text-to-text-generation:program_synthesis}.
He built the platform for the human experiment , publicized the experiment to get votes, and compiled the results for \ref{sct:experiments:negative-prompting}.

\paragraph{Honglu Fan} proofread \ref{sct:methodology:guidance-llm}, \ref{sct:methodology:guidance-t2i}, redacted Section \ref{sct:experiments}'s introduction and \ref{sct:experiments:benchmarks}, Appendix \ref{appendix:charts:general-benchmarks}\ref{app:flops}, \ref{appendix:charts:humaneval}. Designed and ran the experiments for Section \ref{sct:experiments:text-to-text-generation}. He took care of running the experiments of Section \ref{sct:experiments:benchmarks} thanks to his access to CoreWeave and Stability's computing cluster.

\paragraph{Alexander Spangher} proofread the paper and is the primary writer/editor and redactor of it. He wrote the Introduction, Section \ref{sct:methodology}'s introduction, Section
\ref{sct:cost-analysis}, Section \ref{sct:explanation}'s introduction, Appendix \ref{app:related} and the Conclusion, regenerated many of the figures, and proofread everything.

He designed, ran and redacted the experiments in Sections \ref{sct:explanation:entropy}, \ref{sct:explanation:instruction}, and Appendix \ref{app:instruction-tuning-comparison}.

\paragraph{Elad Levi} designed and ran the Chain-Of-Thoughts experiments in Section \ref{sct:experiments:cot}. He wrote a preliminary version of Sections \ref{sct:methodology:guidance-t2i}, \ref{sct:methodology:guidance-llm} and redacted Section \ref{sct:experiments:cot} and Appendix \ref{app:cot}.

\paragraph{Pawan Ammanamanchi} designed, ran and redacted the machine translation experiments of Appendix \ref{app:machine-translation}. 

\paragraph{Stella Biderman} supervised the process. She proofread the paper, suggested the experiments to run in \ref{sct:experiments:benchmarks} and how to run them with EleutherAI's LM Harness. She suggested the GPT-J code generation experiment of section \ref{sct:experiments:text-to-text-generation:program_synthesis}.

\section{Additional Related Works}
\label{app:related}

\subsection{CFG}

The work on CFG is based on Classifier Guided Diffusion \cite{guidance}, which demonstrates that $\gamma$ allows for trading fidelity and diversity. Artists using Stable Diffusion, an open-source product built on \cite{latentdiffusion}, commonly believe that effective prompt engineering and creative pictures require strong prompt conditioning happening for $\gamma>1$. This belief is supported by experiments, such as those conducted with Imagen \cite{imagen}, which show that the prompt correlates more with the image as $\gamma$ increases.

\subsection{Generative Guidance in NLP}

Co-temporaneously with the earliest advances in neural language modeling \cite{gers2000learning} came the recognition that the outputs of these models had to be guided in order to be coherent \cite{barzilay2008modeling} and focused \cite{holtzman2019curious}. And when larger, higher-performing models like GPT \cite{radford2019language, brown2020language} began to show real-world use-cases, the recognition emerged of the need to control their output \cite{solaiman2019release} to guard against toxic content \cite{gehman2020realtoxicityprompts} and bias \cite{felkner2022towards}. 

A central thrust in recent NLP research been to address the above concerns, and approaches have been targeted at nearly every step of training and querying models, from dataset curation \cite{falcon40b} and training \cite{keskar2019ctrl}, to response-alignment \cite{ouyang2022training} and prompt-identification \cite{gehman2020realtoxicityprompts}.

Our work aligns with efforts to control the output of language models by controlling the model's outputted vocabulary distribution $p(x_n | x_{<n})$. Early efforts in this vein aimed at increasing coherence include now-standard techniques like temperature-scaling \cite{chorowski2016towards}, nucleus sampling \cite{holtzman2019curious} and heuristics (e.g. repetition penalties \cite{fu2021theoretical}). 

In parallel, more sophisticated approaches to control the output of language models by moderating the vocabularly distribution emerged within the line of ``controlled text generation''. Works in this vein emerged \textit{after} the earliest attempt at controlled-generation, CTRL \cite{keskar2019ctrl}, where researchers pretrained a language model to be aware of prompts as well as ``control codes'', $a$ that could produce conditional generations, $p(x_n | x_{<n}, a)$, (where $a \in \{$ ``Science'', ``Romance'', ``Mystery''$...\}$) that could produce conditional generations, steer the prompt continuation away from the initial generation. This work established the idea of ``controlled generation''; it was quickly followed by the Plug and Play Language model (PPLM) \cite{dathathri2019plug}. PPLM was the earliest work achieving controlled generation through moderating the vocabulary distribution of a vanilla pretrained language model. Authors used Bayes Rule to factorize the conditional distribution $p(x_n | x_{<n}, a) \propto p(x_n | x_{<n}) p(a | x_n, x_{<n})$. Other works followed in this vein \cite{krause2020gedi, yang2021fudge, spangher2023sequentially, meng2022controllable, li2022diffusion}. Authors used a naive pretrained language model like GPT2 \cite{radford2019language} to model $p(x_n | x_{<n})$ and trained a discriminator $p(a | x)$ on labeled datasets, and then added together the two log probabilities to obtain the controlled distribution.

Efforts at controlled generation largely fell out of favor with the advent of instruction-tuning \cite{ouyang2022training}; using instruction-tuned models like GPT3 \cite{brown2020language}, users could simply the model to ``write happy text'', or ``write very happy text''. However, experiments with moderating the vocabulary distribution continued, and researchers recently showed that combining two models -- an expert model and a weak model -- could produce more fluent text \cite{li2022contrastive}. In this paper, instead of our CFG formulation ($\lambda \log p(x | y) - (1-\lambda) \log p(x)$), authors used two models, a weak model $f_w$ and a strong model $f_s$, to do: $f_s(x|y) - f_w(x|y)$ in order to generate more inventive, creative language that was even \textit{more} in the direction of $f_s$ than would have been.

\section{Charts}
In this section, we collect some charts that visualize results in Section \ref{sct:experiments:benchmarks}, \ref{sct:experiments:text-to-text-generation} and \ref{sct:explanation}.

\subsection{General benchmarks}\label{appendix:charts:general-benchmarks}
\begin{figure}
    \centering
    \includegraphics[width=\textwidth]{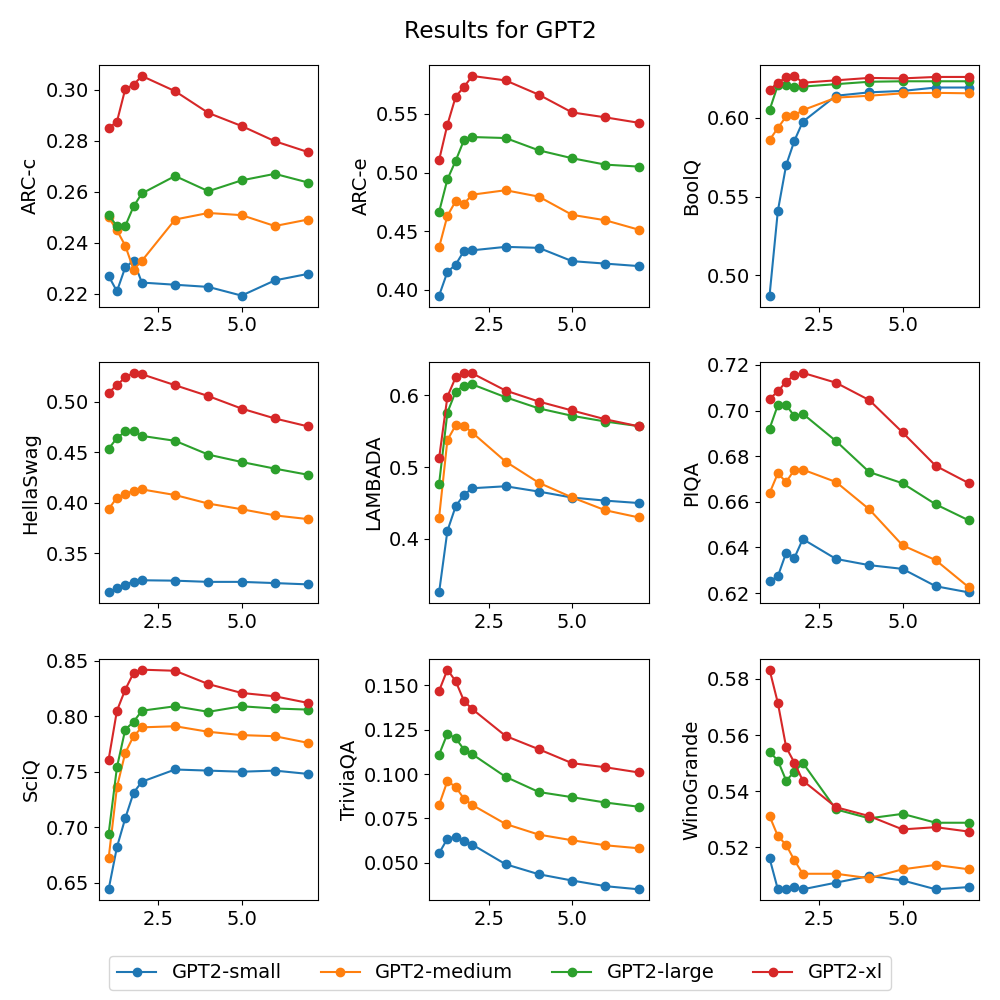}
    \caption{Standard benchmarks over various CFG strengths for GPT2 models}
    \label{fig:gpt2-split}
\end{figure}

\begin{figure}
    \centering
    \includegraphics[width=\textwidth]{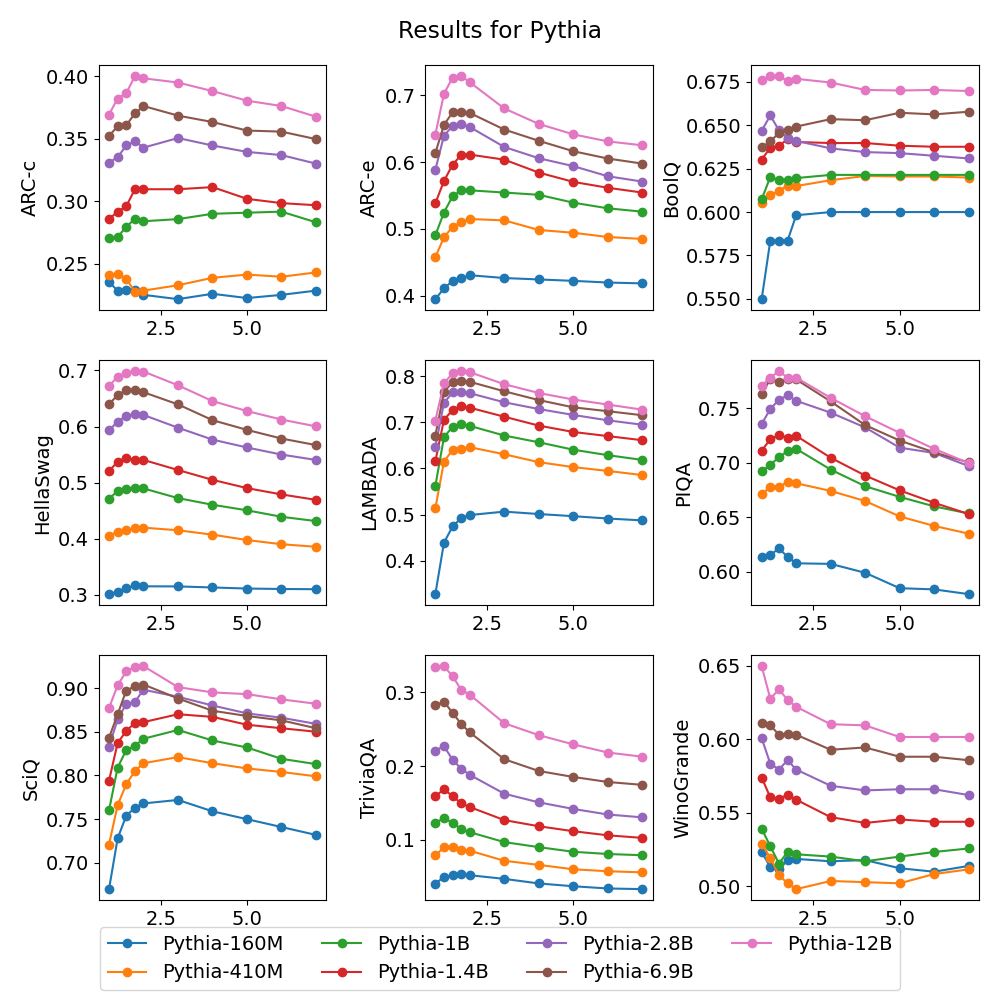}
    \caption{Standard benchmarks over various CFG strengths for Pythia models}
    \label{fig:pythia-split}
\end{figure}

\begin{figure}
    \centering
    \includegraphics[width=\textwidth]{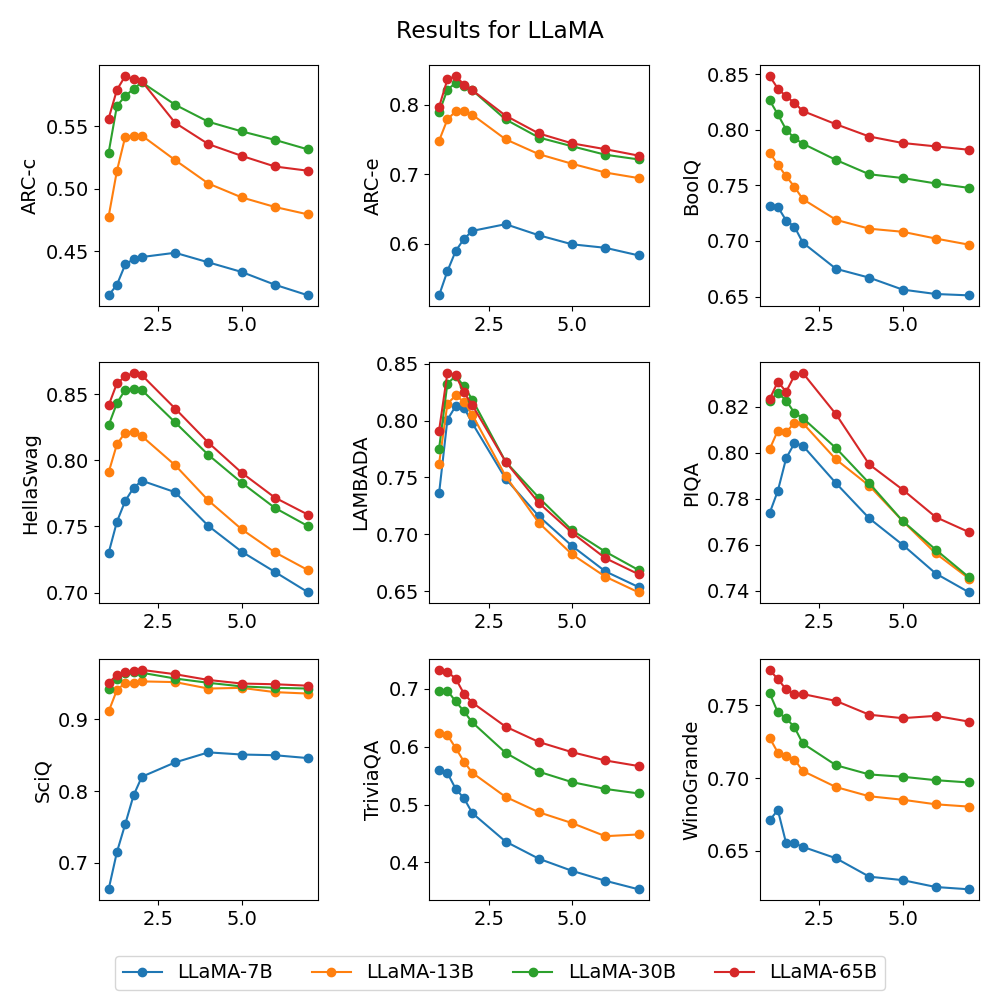}
    \caption{Standard benchmarks over various CFG strengths for LLaMA models}
    \label{fig:llama-split}
\end{figure}

In Section \ref{sct:experiments:benchmarks}, GPT-2, Pythia, LLaMA model families are analyzed with and without CFG. In addition to Table \ref{tab:benchmark}, we make plots of each model family with x-axis being the CFG strength and the y-axis being the accuracy. It aims to provide a more direct view of how model size affect the accuracy-to-$\gamma$ curves while scaling in the same model family. The plots are shown in Figure \ref{fig:gpt2-split}, \ref{fig:pythia-split} and \ref{fig:llama-split}.

We run TriviaQA based on the LLaMA \cite{llama} methodology, however we perform substring match rather than exact match. This stems from manual analysis which showed that exact matching disqualified answers like \texttt{"Mark Twain"} (with quotes) or \texttt{His name is Mark Twain} instead of the exact \texttt{Mark Twain}.

\subsection{Accuracy vs. FLOP}
\label{app:flops}
In Section \ref{sct:cost-analysis}, we present the finding that a model using CFG can generally perform as well as a model twice as large without CFG. The detailed charts are presented in this subsection.

\begin{figure}
    \centering
    \includegraphics[width=\textwidth]{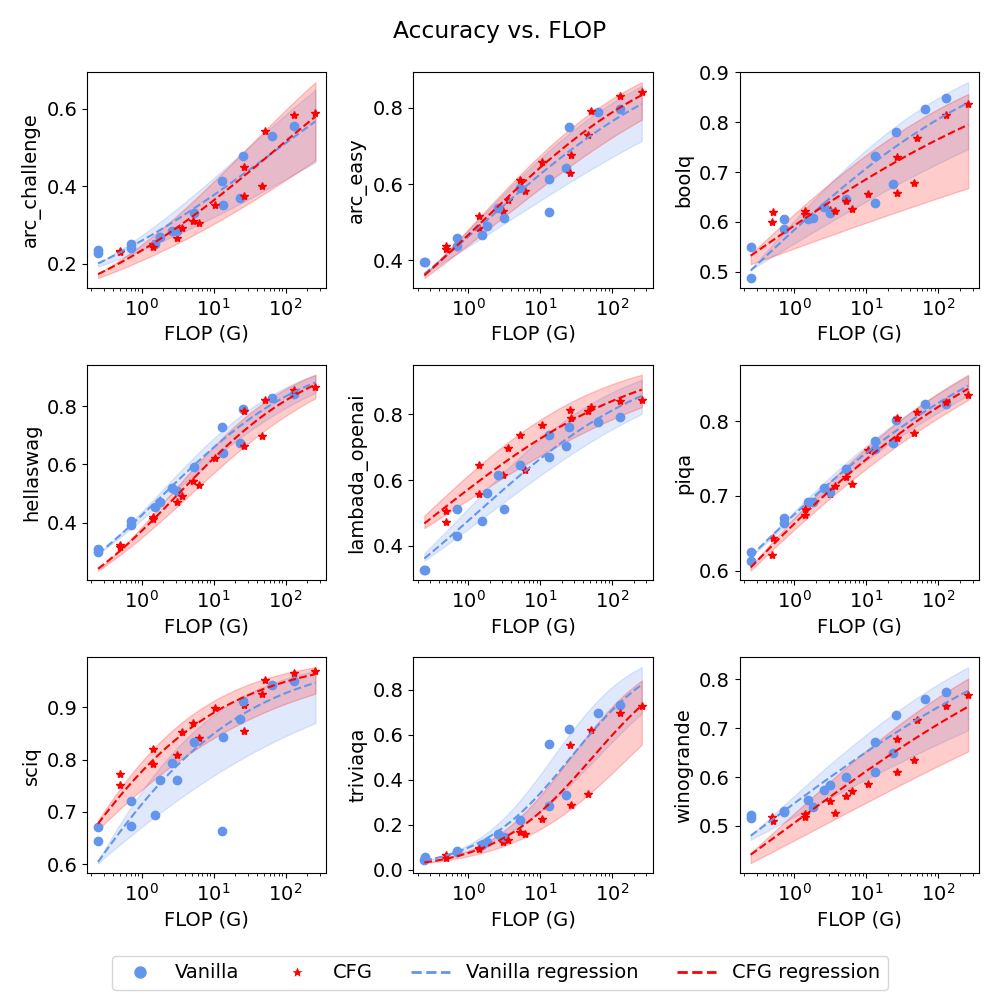} .
    \caption{Accuracy vs. FLOP per token at inference. \\\textbf{Blue point}: a model without CFG from any of the three model families (GPT-2, Pythia, LLaMA). \\\textbf{Red point}: a model with the best CFG from any of the three model families. \\\textbf{The dashed curves}: the regression curves (logistic regression between log-FLOP and accuracy) of their groups.
    }
    \label{fig:flops}
\end{figure}

With the same data points as Section \ref{appendix:charts:general-benchmarks}, we reorganize them into inference accuracy vs. FLOP\footnote{FLOP: floating point operations} per token plots so that we can compare the performance of a model with CFG (doubled inference FLOP) and a model without CFG but twice as big. We show all the plots in Figure \ref{fig:flops}.

Note that:
\begin{enumerate}
    \item \textbf{The location of each data point in the charts ignores the model size and only reflects its inference FLOP per token.} For example, a 1.4B model with CFG (doubled inference FLOP) will show up near a 2.8B model without CFG if they perform closely, despite the fact that such 1.4B model is more useful in practice due to the saving on training and VRAM.
    \item \textbf{The data points in the charts only reflect the inference cost and ignoring the training cost.} For example, when a 1.4B model gets boosted to the accuracy of a 2.8B model by using CFG, the inference costs are similar but to train a 1.4B model takes less compute.
\end{enumerate}

For Lambada and SciQ, CFG is a clear winner which improves the whole compute-accuracy curve while for WinoGrande, CFG impacts negatively. The rest are mixed.

This entails that for the same inference cost, CFG can emulate a model that has twice the parameter count. This drastically reduces the VRAM usage needed to run the models which is the current bottleneck, and reduces the training cost. To further justify this, Table \ref{fig:flops} is a breakdown of the ANCOVA $p$-values for each chart between the regression line of the CFG group (in red) and the one of the vanilla group (in blue). We choose the $p$-value cutoff at $0.01$ according to \cite{rutherford2011anova}, and higher than $0.01$ means an insignificant difference between the regression lines of the two groups.

\begin{table}
\centering
\begin{tabular}{lrr}
\toprule
 & $p$-value & Win \\
\midrule
Lambada & 0.000 & CFG \\
WinoGrande & 0.003 & Vanilla \\
SciQ & 0.008 & CFG \\
TriviaQA & 0.008 & Vanilla \\
HellaSwag & 0.012 &  $p>.01$\\
PiQA & 0.030 &  $p>.01$\\
ARC-c & 0.216 & $p>.01$ \\
BoolQ & 0.345 & $p>.01$ \\
ARC-e & 0.355 & $p>.01$ \\
\bottomrule
\end{tabular}
\caption{ANCOVA $p$-value results for plots shown in Figure \ref{fig:flops}. We calculate ANCOVA on log-transformed variables and calculate significance at $p=.01$.}
\end{table}

\subsection{HumanEval benchmark}\label{appendix:charts:humaneval}
\begin{figure}[!ht]
    \includegraphics[width=\linewidth]{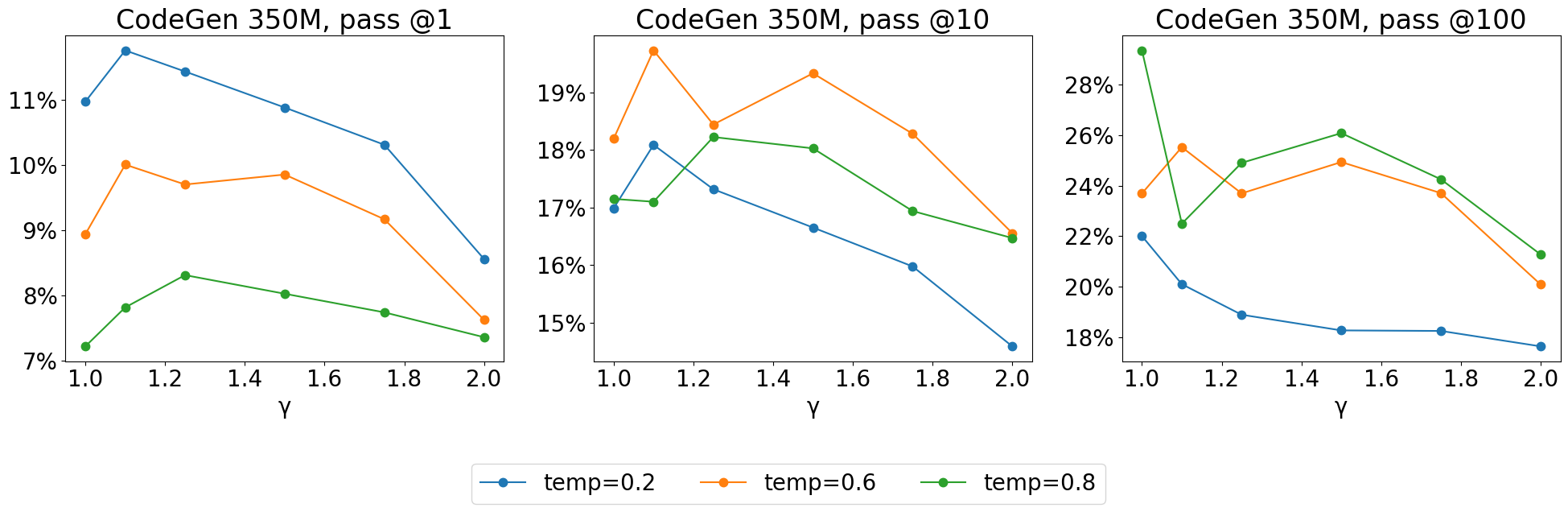}
    \caption{CodeGen-350M-mono performance on HumanEval with various CFG strengths}\label{fig:codegen350}
\end{figure}

\begin{figure}[!ht]
    \includegraphics[width=\linewidth]{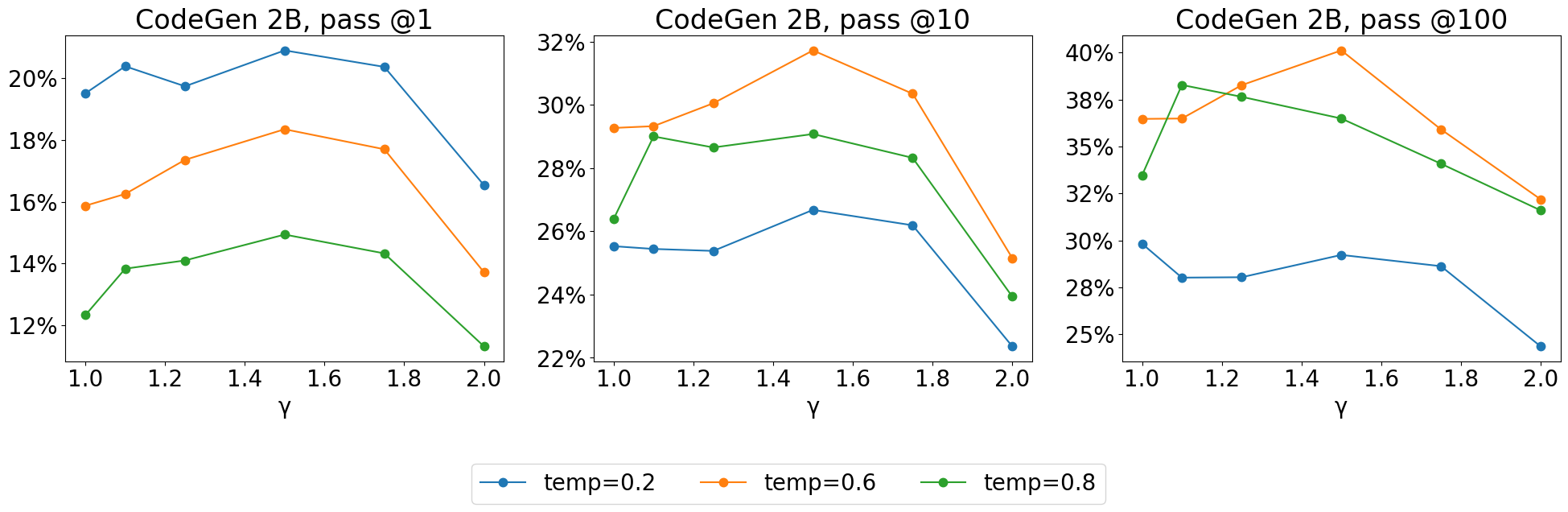}
    \caption{CodeGen-2B-mono performance on HumanEval with various CFG strengths}\label{fig:codegen2b}
\end{figure}

\begin{figure}[!ht]
    \includegraphics[width=\linewidth]{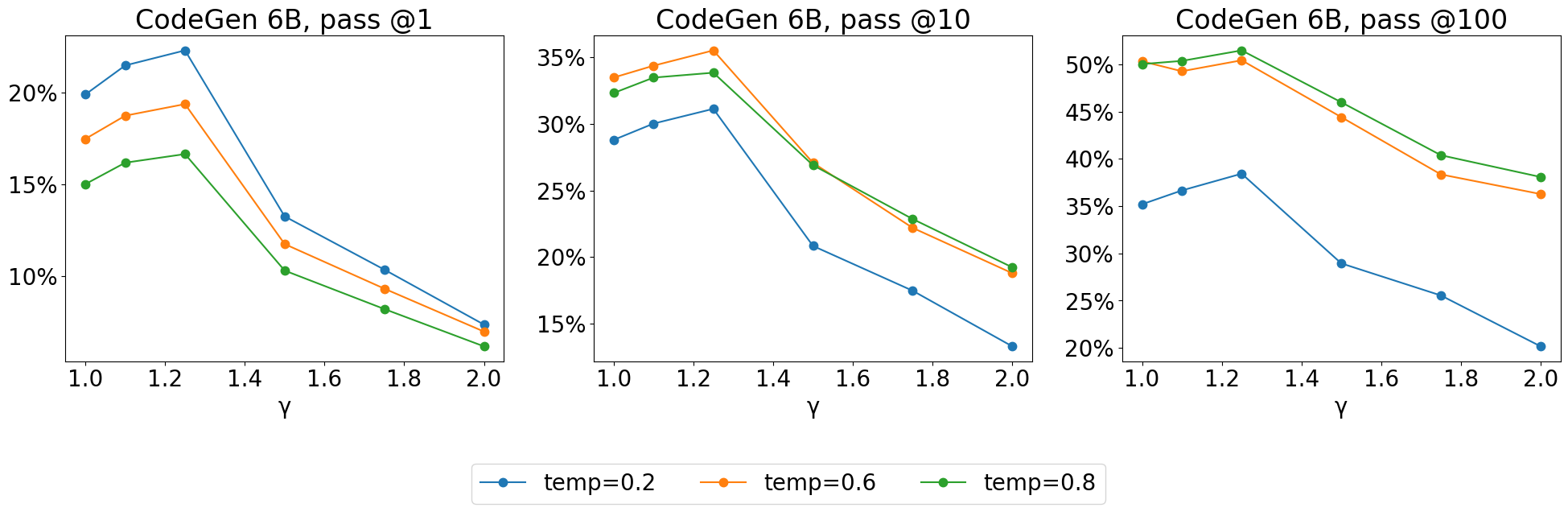}
    \caption{CodeGen-6B-mono performance on HumanEval with various CFG strengths}\label{fig:codegen6b}
\end{figure}

In Section \ref{sct:experiments:text-to-text-generation:program_synthesis}, we explain our experiments on CodeGen-350M-mono, CodeGen-2B-mono and CodeGen-6B-mono and show their performances in the HumanEval benchmark with various CFG for temperature $0.2$ in Table \ref{tab:codegen}. The full results for temperature $=0.2, 0.6, 0.8$ are shown below in Table \ref{tab:codegen-350}, \ref{tab:codegen-2b} and \ref{tab:codegen-6b}). We also put the pass@$k$-to-$\gamma$ curves of different temperatures together to show how the temperatures affect the impact of CFG when the model size and $k$ are fixed in Figure \ref{fig:codegen350}, \ref{fig:codegen2b} and \ref{fig:codegen6b}.

\begin{table}[t]
    \centering
    \begin{tabular}{llllllllll}
        \toprule
         & \multicolumn{3}{c}{temperature = 0.2} & \multicolumn{3}{c}{temperature = 0.6} & \multicolumn{3}{c}{temperature = 0.8} \\
         \cmidrule(r){2-4} \cmidrule(r){5-7} \cmidrule(r){8-10}
        $\gamma$ & k=1 & k=10 & k=100 & k=1 & k=10 & k=100 & k=1 & k=10 & k=100 \\
        \midrule
        1.0 & 11.0\% & 17.0\% & \textbf{22.0\%} & 8.9\% & 18.2\% & 23.7\% & 7.2\% & 17.2\% & \textbf{29.4\%} \\
        1.1 & \textbf{11.8\%} & \textbf{18.1\%} & 20.1\% & \textbf{10.0\%} & \textbf{19.7\%} & \textbf{25.5\%} & 7.8\% & 17.1\% & 22.5\% \\
        1.25 & 11.4\% & 17.3\% & 18.9\% & 9.7\% & 18.4\% & 23.7\% & \textbf{8.3\%} & \textbf{18.2\%} & 24.9\% \\
        1.5 & 10.9\% & 16.7\% & 18.3\% & 9.9\% & 19.3\% & 24.9\% & 8.0\% & 18.0\% & 26.1\% \\
        1.75 & 10.3\% & 16.0\% & 18.2\% & 9.2\% & 18.3\% & 23.7\% & 7.7\% & 16.9\% & 24.2\% \\
        2.0 & 8.6\% & 14.6\% & 17.6\% & 7.6\% & 16.6\% & 20.1\% & 7.4\% & 16.5\% & 21.3\% \\
        \bottomrule
    \end{tabular}
    \caption{CodeGen-350M-mono results}
    \label{tab:codegen-350}
\end{table}

\begin{table}[t]
    \centering
    \begin{tabular}{llllllllll}
        \toprule
         & \multicolumn{3}{c}{temperature = 0.2} & \multicolumn{3}{c}{temperature = 0.6} & \multicolumn{3}{c}{temperature = 0.8} \\
         \cmidrule(r){2-4} \cmidrule(r){5-7} \cmidrule(r){8-10}
        $\gamma$ & k=1 & k=10 & k=100 & k=1 & k=10 & k=100 & k=1 & k=10 & k=100 \\
        \midrule
        1.0 & 19.5\% & 25.5\% & \textbf{29.8\%} & 15.9\% & 29.3\% & 36.5\% & 12.3\% & 26.4\% & 33.5\% \\
        1.1 & 20.4\% & 25.4\% & 28.0\% & 16.3\% & 29.3\% & 36.5\% & 13.8\% & 29.0\% & \textbf{38.3\%} \\
        1.25 & 19.7\% & 25.4\% & 28.0\% & 17.4\% & 30.1\% & 38.3\% & 14.1\% & 28.7\% & 37.6\% \\
        1.5 & \textbf{20.9\%} & \textbf{26.7\%} & 29.2\% & \textbf{18.3\%} & \textbf{31.7\%} & \textbf{40.1\%} & \textbf{14.9\%} & \textbf{29.1\%} & 36.5\% \\
        1.75 & 20.4\% & 26.2\% & 28.6\% & 17.7\% & 30.4\% & 35.9\% & 14.3\% & 28.3\% & 34.1\% \\
        2.0 & 16.5\% & 22.4\% & 24.4\% & 13.7\% & 25.2\% & 32.2\% & 11.3\% & 23.9\% & 31.6\% \\
        \bottomrule
    \end{tabular}
    \caption{CodeGen-2B-mono results}
    \label{tab:codegen-2b}
\end{table}

\begin{table}[t]
    \centering
    \begin{tabular}{llllllllll}
        \toprule
        & \multicolumn{3}{c}{temperature = 0.2} & \multicolumn{3}{c}{temperature = 0.6} & \multicolumn{3}{c}{temperature = 0.8} \\
         \cmidrule(r){2-4} \cmidrule(r){5-7} \cmidrule(r){8-10}
        $\gamma$ & k=1 & k=10 & k=100 & k=1 & k=10 & k=100 & k=1 & k=10 & k=100 \\
        \midrule
        1.0 & 19.5\% & 25.5\% & \textbf{29.8\%} & 15.9\% & 29.3\% & 36.5\% & 12.3\% & 26.4\% & 33.5\% \\
        1.1 & 20.4\% & 25.4\% & 28.0\% & 16.3\% & 29.3\% & 36.5\% & 13.8\% & 29.0\% & \textbf{38.3}\% \\
        1.25 & 19.7\% & 25.4\% & 28.0\% & 17.4\% & 30.1\% & 38.3\% & 14.1\% & 28.7\% & 37.6\% \\
        1.5 & \textbf{20.9\%} & \textbf{26.7\%} & 29.2\% & \textbf{18.3\%} & \textbf{31.7\%} & \textbf{40.1\%} & \textbf{14.9\%} & \textbf{29.1\%} & 36.5\% \\
        1.75 & 20.4\% & 26.2\% & 28.6\% & 17.7\% & 30.4\% & 35.9\% & 14.3\% & 28.3\% & 34.1\% \\
        2.0 & 16.5\% & 22.4\% & 24.4\% & 13.7\% & 25.2\% & 32.2\% & 11.3\% & 23.9\% & 31.6\% \\
        \bottomrule
    \end{tabular}
    \caption{CodeGen-6B-mono results}
    \label{tab:codegen-6b}
\end{table}

\begin{figure}[!ht]
    \centering
    \subfloat[CodeGen-350M-mono HumanEval task-by-task plot with temp=$0.8$ \\
    \textbf{Blue}: CFG outperforms,\\ \textbf{Purple}: CFG ties with the baseline,\\\textbf{Red}: CFG underperforms]{
    \includegraphics[width=.5\linewidth]{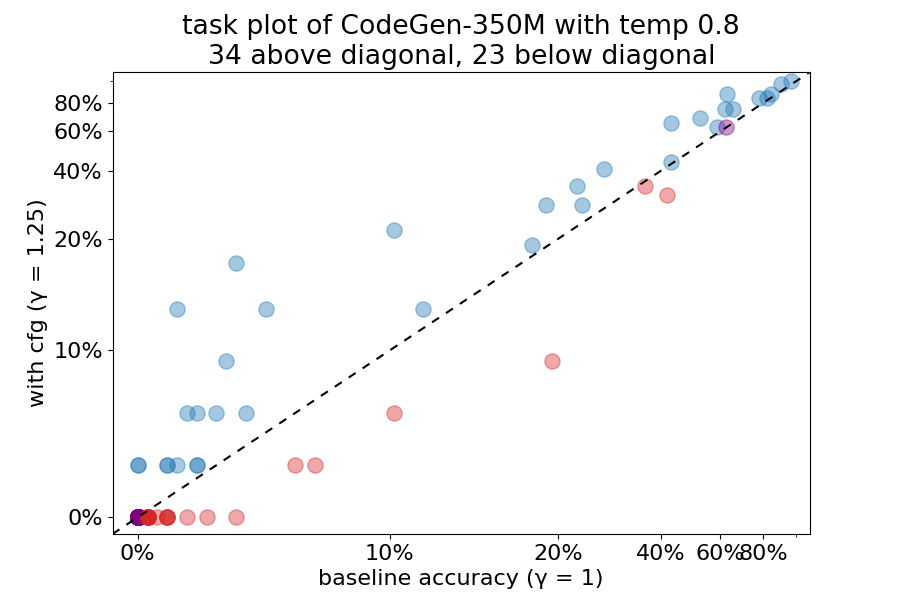} 
    \label{fig:humaneval-tasks-0.8}
    }
    \subfloat[CodeGen-350M-mono HumanEval task-by-task plot with temp=$0.6$ \\
    \textbf{Blue}: CFG outperforms,\\\textbf{Purple}: CFG ties with the baseline,\\\textbf{Red}: CFG underperforms]{\includegraphics[width=.5\linewidth]{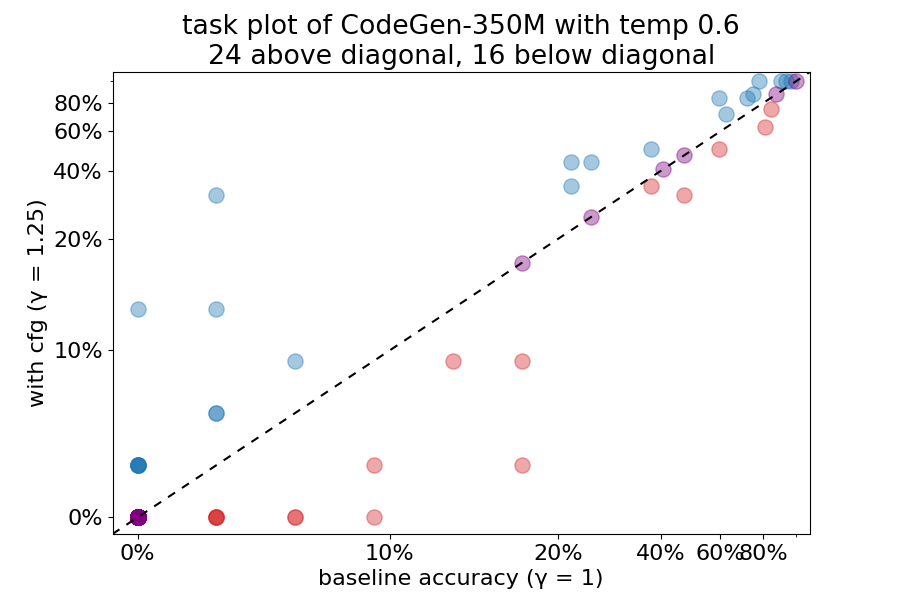}
    \label{fig:humaneval-tasks-0.6}
    } \\ 
    \subfloat[CodeGen-350M-mono HumanEval task-by-task plot with temp=$0.2$ \\
    \textbf{Blue}: CFG outperforms,\\\textbf{Purple}: CFG ties with the baseline,\\\textbf{Red}: CFG underperforms]{\includegraphics[width=.5\linewidth]{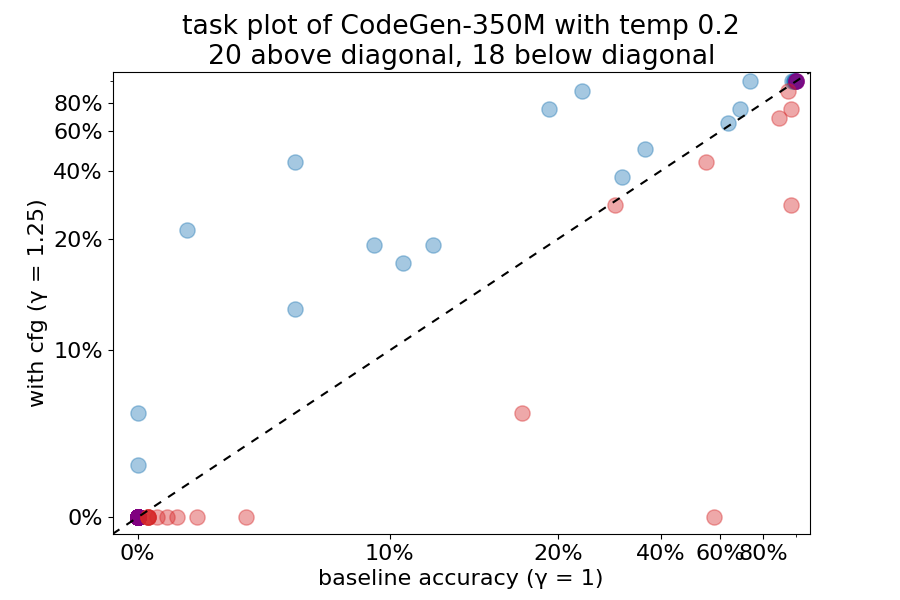}
    \label{fig:humaneval-tasks-0.2}
    }
\end{figure}

In addition, we breakdown the result of CodeGen-350M-mono on HumanEval benchmark into individual tasks. We plot the ``accuracy with cfg" vs. ``accuracy without cfg" charts to visualize the outperform/underperform distributions among all tasks. The plots are shown in Figure \ref{fig:humaneval-tasks-0.2}, \ref{fig:humaneval-tasks-0.6} and \ref{fig:humaneval-tasks-0.8}.

\subsection{Deliberative Prompting: Chain-of-Thought}
\label{app:cot}
\begin{figure}[!ht]
    \includegraphics[width=.4\linewidth]{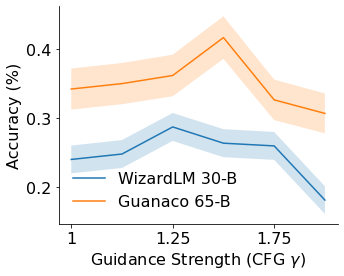}
    \includegraphics[width=.4\linewidth]{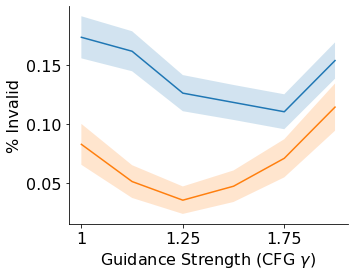}
    \caption{CFG impact on chain-of-thought prompting with respect to AQuA dataset.  For small CFG values, using CFG increases the percentage of chains which end in a valid answer structure while increasing the model accuracy. For large values the invalid percentage remains small but the accuracy drop.}\label{fig:cot_aqua}
\end{figure}

In this subsection we provide additional results for \ref{sct:experiments:cot}. In Figure~\ref{fig:cot_aqua} we provide results on AQuA dataset and in Tables~\ref{tbl:cot_qual1} and ~\ref{tbl:cot_qual2} we provide a qualitative comparison of CoT with and without CFG. These results support our finding that using CFG increases the percentage of CoT which results
in a valid answer and boost the model performances. 

\begin{figure}[t]
    \centering
    \includegraphics[width=\linewidth]{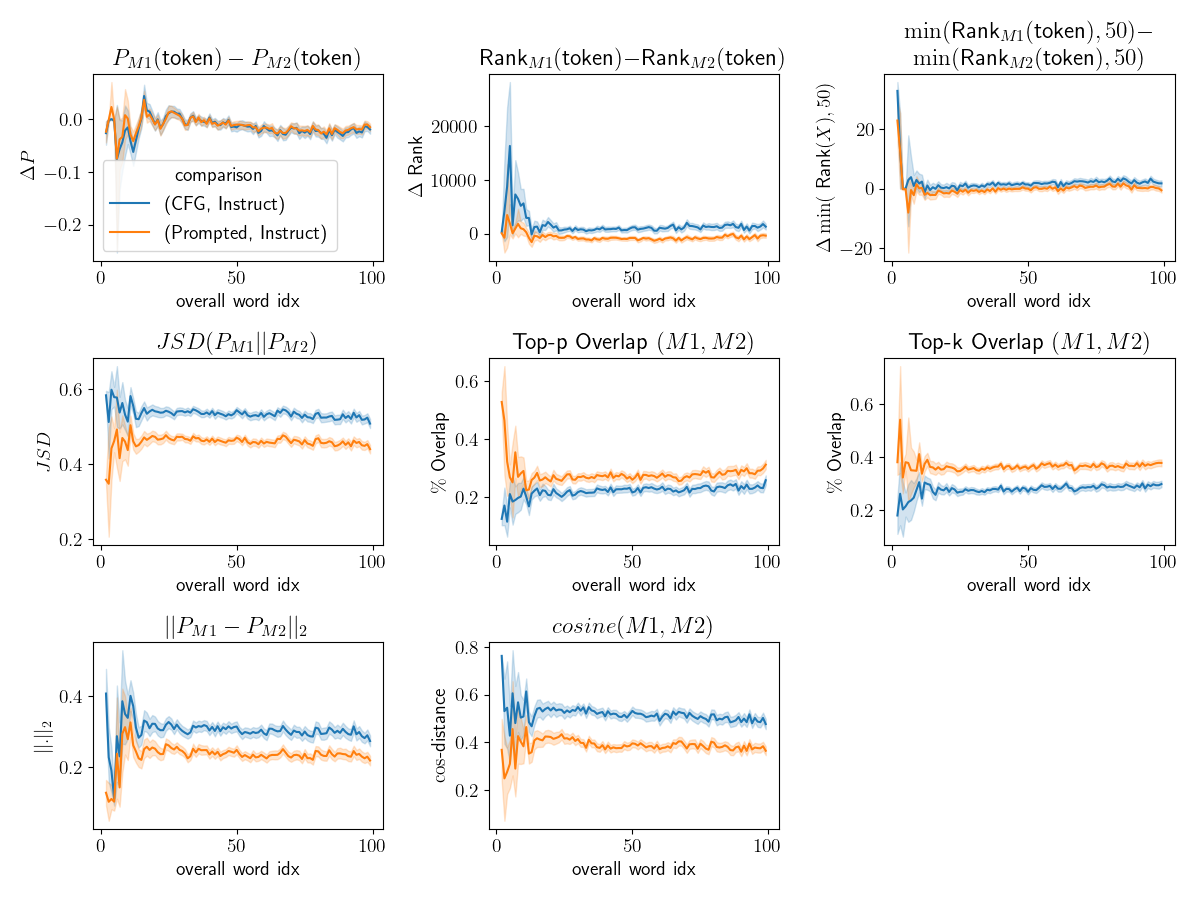}
    \caption{Comparison of (CFG-$\gamma=1.5$, Instruct) logits across a large sample set from P3.}
    \label{fig:instruct-cfg-comparison-subplots}
\end{figure}

\begin{table}[t]
\centering
\begin{tabular}{lrlr}
\toprule
P3 Dataset &   mean &       std &  count \\
\midrule
Highest $\langle$ CFG, Instruct$\rangle$ Similarities&        &           &        \\
\midrule
SuperGLUE wsc.fixed p is are r score eval        &  31.89 &  +/-22.06 &     42 \\
SciQ Multiple Choice Closed Book                  &   5.82 &  +/-13.27 &     43 \\
CosE v1.11 description question option text      &   5.70 &   +/-9.05 &     43 \\
RottenTomatoes Writer Expressed Sentiment        &   4.93 &   +/-7.45 &     41 \\
WinograndeXL fill in the blank        &   4.42 &  +/-10.51 &     44 \\
RottenTomatoes Text Expressed Sentiment          &   2.93 &   +/-7.98 &     45 \\
Quarel: choose between                             &   2.51 &  +/-12.39 &     43 \\
SuperGLUE wic GPT 3 prompt score eval            &   2.15 &   +/-5.94 &     44 \\
WinograndeDebiased Replace score eval &   2.02 &  +/-24.46 &     41 \\
PAWS final context question (no label)      &   1.37 &   +/-4.81 &     43 \\
\midrule
Lowest $\langle$ CFG, Instruct$\rangle$ Similarities&        &           &        \\
\midrule
paws labeled final paraphrase task                & -11.71 &  +/-11.03 &     42 \\
super glue copa more likely                       & -11.94 &   +/-6.38 &     45 \\
piqa Does this solution make sense sol2           & -12.22 &   +/-9.24 &     42 \\
super glue copa cause effect score eval           & -12.82 &    +/-5.8 &     41 \\
rotten tomatoes Sentiment with choices            & -13.07 &   +/-7.96 &     41 \\
super glue copa plausible alternatives score eval & -15.07 &   +/-5.69 &     41 \\
super glue copa C1 or C2 premise so because       & -15.38 &   +/-6.43 &     41 \\
super glue copa more likely score eval            & -16.54 &   +/-5.45 &     43 \\
cos e v1.11 question option description id        & -17.60 &  +/-14.06 &     41 \\
rotten tomatoes Reviewer Enjoyment Yes No         & -18.16 &  +/-16.02 &     45 \\
\bottomrule
\end{tabular}
\caption{Datasets in P3 where Instruction-Tuned models were the most and least similar, in terms of top-p overlap, to CFG models. The count column shows the number of datapoints that were sampled from each dataset to calculate the overlap.}
\label{tab:cfg_instruct_datasets}
\end{table}

\begin{table}[t]
    \centering
    \begin{tabularx}{\linewidth}{>{\hangindent=2em}X}
    \toprule
    Top Sentences in P3 where CFG is MOST Similar to Instruction-Tuned Models\\
    \midrule
Build a movie plot around this: What is the team? Rag-tag bunch of girls\\
Here's a complex question that requires someone to reason about the input, can you answer it? What city was the capital of the Ostrogothic Kingdom and the birth place of Ornella Fiorentini?\\
Who had more of their English novels turned into Oscar-nominated films, Raja Rao or Pat Conroy?\\
Nokia, Texas Instruments and other leading makers of mobile phones have formally complained to Brussels that Qualcomm, the US mobile chipmaker, has unfairly used its patents on 3G technologies. Question: Texas Instruments produces mobile phones. True or False?\\
Context: Patting her back, the woman smiled at the girl .    Question: "her" is the woman. True or false?   Answer:\\
Take the following as truth: The American Combat Association is a small mixed martial arts company founded by Olympic wrestler, world Abu Dhabi champion and UFC fighter Kamal Shalorus and professional mixed martial arts fighter, Broadcaster and American professional wrestler Matthew "The Granimal" Granahan.  Then the following statement: "The American Combat Association was founded by two Olympic wrestlers." is true, false, or inconclusive? \\
Pick the most correct option to answer the following question.  Some antibiotics used to treat infections in humans are also used to treat chickens, but some groups oppose this practice. The overuse of the antibiotics will most likely influence the natural selection of which type of organisms?  Options:  - A: chickens that naturally make the antibiotics  - B: microbes that are resistant to the antibiotics  - C: microbes that are susceptible to the antibiotics  - D: chickens that are resistant to infection  \\
Jennifer dragged Felicia along to a self help workshop about how to succeed, because \_ wanted some company. Replace the \_ in the above sentence with the correct option:  - Jennifer - Felicia \\
Brian could learn to swim with the right instruction, but it was hard to tell whether lifeguard Matthew was qualified to provide it, since \_ had never swum before. Replace the \_ in the above sentence with the correct option:  - Brian - Matthew \\
\bottomrule
\end{tabularx}
    \caption{Top sentences in P3 where CFG is similar to Instruction-Tuned models, as measured by top-$p$ overlap.}
    \label{tab:topp_cfg}
\end{table}

\begin{table}[t]
    \centering
    \begin{tabularx}{\linewidth}{>{\hangindent=2em}X}
    \toprule
    Sentences in P3 where CFG is LEAST Similar to Instruction-Tuned Models\\
    \midrule
How do you feel about your current weight and eating habits ?\\
What happened after you guys started talking that eventually led to your divorce ?\\
Given a goal and a wrong solution, rewrite it to give a correct solution. Goal: how do you train a puppy?  Solution:  Corrected solution:\\
What might have happened since I was a democrat in my first year ?\\
What do you usually do when you meet a guy for the first time ?\\
What did you do that caused you to be in the bathroom all day ?\\
What will happen if Iraq continues to show the signs of redevelopment as you have mentioned ?\\
What might happen if we show our true selves to the people we love ?\\
I would like to create a garden on my balcony. What is the first thing I should do?\\
What will you do if a branch falls off one of the oaks ?\\
What will you do now that you define as taking action ?\\
The abode of the Greek gods was on the summit of Mount Olympus, in Thessaly. Question: Mount Olympus is in Thessaly. True or False?\\
Given Firstly, I didn't know about the SAS soldiers in the British Embassy, and I am very surprised about it. Very surprised indeed, Ambassador. Secondly I do not think it is a good idea to attack a plane with a hundred and seven passengers in it and ``take it apart'' as you say. Is it guaranteed true that "it is a good idea to attack a plane with a hundred and seven passengers in it and 'take it apart'"? Yes, no, or maybe?\\
'Cote d'Ivoire's President, Laurent Gbagbo, promulgated new election laws on July 14. Question: President Laurent Gbagbo lives in Cote d'Ivoire. True or False?\\
'the real star of this movie is the score , as in the songs translate well to film , and it's really well directed . The sentiment expressed for the movie is '\\
My closet was messy.  so...  Choose between: - I organized it. - I decorated it.\\
\bottomrule
\end{tabularx}
    \caption{Sentences in P3 where CFG is LEAST similar to Instruction-Tuned models, as measured by top-$p$ overlap.}
    \label{tab:bottom_p_instruct}
\end{table}

\FloatBarrier

\section{Additional experiments}

\subsection{Machine translation}
\label{app:machine-translation}

We evaluate using Classifier-Free Guidance for machine translation on a variety of models. We choose the WMT14 fr-en \cite{bojar-EtAl:2014:W14-33_wmt14_fr_en} as the dataset of choice to understand if CFG would also help multilingual datasets. We run 0-shot experiments on Bloom-3B \cite{Scao2022BLOOMA1}, a multilingual model trained on 49 languages. We also test on RedPajama-Incite-Base-3B, trained on 1.5T tokens of English text and mT0 \cite{Muennighoff2022CrosslingualGT_mt0} a prompt tuned sequence-to-sequence model. For the Bloom-3B model, we test for multiple prompts and perform 1-shot experiments as well.  All scores are measured in BLEU.

We find that for this generation task, $\gamma$ ranging between 1.1 to 1.25 yield the best results and perform increasingly worse at higher values. We additionally observe that the method is prompt-invariant, showing gains regardless of the prompt choice in 0-shot performance. We do not see any improvements in the case of 1-shot performance for Bloom-3B. We also do not see any significant performance gains in the case of mT0, suggesting that prompt-tuned models might already be at the pinnacle of possible 0-shot performance.

\begin{table}[t]
    \centering
    \begin{tabular}{lccr} 
    \toprule     Model    & $\gamma=1$ & $\gamma=1.10$ & $\gamma=1.25$ \\
         \midrule
        Bloom-3B  & $14.16$ & \textbf{15.81} & $14.16$ \\
        RedPajama-Incite-3B & $15.04$ & \textbf{17.24} & $17.78$ \\
        \midrule
                         & $\gamma=1$ & $\gamma=1.05$ & $\gamma=1.10$ \\
        \midrule
        Bloom-3B 1-shot & \textbf{29.84} & $29.19$ & $28.53$ \\
        mT0  & \textbf{29.77} & \textbf{$29.41$} & $27.79$ \\
        
        \bottomrule
    \end{tabular}
    \caption{BLEU scores for different $\gamma$ for machine translation tasks. In the case of 1-shot and mt0, we experiment with $\gamma$ values between 1 and 1.1 since we see a rapid decline at even slightly higher values. All models are evaluated 0-shot unless otherwise specified.}
    \label{tab:mt-gen}
\end{table}

\subsection{Prompting experiments for code generations}\label{appendix:additional-exp}
We summarize two exploratory experiments which are briefly mentioned in \ref{sct:experiments:text-to-text-generation:program_synthesis} and precedes our systematic evaluations on HumanEval. 
\begin{enumerate}
    \item The first experiment is to prompt GPT-J \cite{gpt-j}\footnote{GPT-J is not specifically trained for code generation task. But it was exposed to some code data in its training.} for code completions of certain languages, and analyze the consistencies between the prompt languages and the completion languages.
    \item The second experiment is to prompt CodeGen-350M-mono \cite{nijkamp2023codegen} to complete a specific image generation function, and analyze multiple aspects of the completions (syntax, the return type, the return shape and the return quality).
\end{enumerate}

Prompting GPT-J for different coding language is inspired by one of the experiments in \cite{moss}. Their observation is that the model often generates non-code or not the programming language it was prompted for.

We generate 100 samples (5 runs for 5 prompts) for each guidance strength $\gamma =  1, 1.25, 1.5, 1.75$. We observe the $\gamma=1$ baseline generating the correct programming language 73\% of the time, jumping to 86\% with $\gamma=1.25$ ($p$-value 0.01). See \ref{table:code-cm} for more details.

\begin{table}
  \centering
  \begin{tabular}{rrrrr|rrrrr}
    \toprule
     $\gamma=1$ & not code & C & Java & Python & $\gamma=1.25$ & not code &  C  & Java & Python \\
     \midrule
Unspecified    &        9 &     9 &    6 &      1 & Unspecified    &        4 &    11 &    9 &      1 \\
C    &        3 &    19 &    3 &      0 & C    &        4 &    19 &    2 &      0 \\
Java &        5 &     0 &   19 &      1 & Java &        2 &     0 &   23 &      0 \\
Python   &        6 &     0 &    0 &     19 & Python   &        1 &     0 &    1 &     23 \\
    \toprule
    $\gamma=1.5$& not code & C & Java & Python & $\gamma=1.75$ & not code &  C  & Java & Python \\
    \midrule
Unspecified & 6 &     8 &    8 &      2 & Unspecified    &        6 &     6 &   10 &      1 \\
C    &        5 &    18 &    2 &      0 & C    &        8 &    16 &    1 &      0 \\
Java &        3 &     0 &   22 &      0 & Java &        2 &     0 &   23 &      0 \\
Python   &        3 &     0 &    0 &     22 & Python   &        5 &     0 &    1 &     19 \\
     
    \bottomrule
  \end{tabular}
  \caption{Confusion matrix for generating code tests with GPT-J. We prompt it to generate code in some programming language (rows) and compare with the generated programming language (columns). The overall accuracy results for $\gamma=1, 1.25, 1.5, 1.75$ are 73\%, 86\%, 81\%, 77\%, respectively. }
  \label{table:code-cm}
\end{table}

Next, we turn to CodeGen-350M-mono (\cite{nijkamp2023codegen}) for code completion for a fixed image generation function. The prompt is the following:
\begin{lstlisting}
# Return a red square on a 32x32 picture in the form of numpy array with RGB channels
def draw() -> np.ndarray:
\end{lstlisting}

We produce $1600$ completions for each CFG strength $\gamma = 1.0, 2.0$. The results are evaluated based on:
\begin{itemize}
    \item syntax correctness (executing without errors),
    \item return type correctness (returning a numpy array),
    \item return shape correctness (having shape $(32, 32, 3)$),
    \item the $l^2$-distance to a reference picture (picture of pure color in red).
\end{itemize}
When calculating the $l^2$-distance, all pixels are normalized to the range $[0, 1]$. The result is summarized in Table \ref{tab:image-gen}.
\begin{table}
    \centering
    \begin{tabular}{lccr} 
    \toprule         & $\gamma=1$ & $\gamma=2$ & improvement \\
         \midrule
        correct syntax & $419$ & $574$ &  137\%\\
        correct return type & $289$ & $546$ & 189\% \\
        correct shape & $284$ & $538$ & 189\% \\
        $l^2$-distance to reference & $0.111$ & $0.090$ & 123\% \\
        \bottomrule
    \end{tabular}
    \caption{Different $\gamma$ for code completion performing the image generation task}
    \label{tab:image-gen}
\end{table}

The difference is fairly noticeable, where the biggest improvement comes from the return type correctness.

\FloatBarrier

\section{Generation samples}
\label{app:samples}

\subsection{Continuations}

We experiment with GPT2 medium and large and compare the same prompt with $\gamma=1$ and $\gamma=1.5$. We show that $\gamma=1.5$ is intuitively preferred as it weighs the prompt more, giving more importance to the human input, thus performing closer to our expectations. Indeed, GPT2-medium[$\gamma=1.5$] gives more satisfactory results than GPT2-large[$\gamma=1$]. These results are not to take lightly: on our CPU, running two GPT2-medium passes per has the same latency of a single GPT2-large pass; GPT2-large is twice as long to evaluate as GPT2-medium and requires twice the the amount of memory, takes more compute to train, and does not expose a controllable prompt conditioning, while being less satisfactory.

\begin{figure}
    \begin{quote}
\textbf{The dragon flew over Paris, France}, and landed in the courtyard of the French parliament, where it was promptly seized by the French government.

But the scene was not quite what it seemed.

The dragon that had been brought here was actually a puppet of the Nazi government, which had been working under the cover of the German government to make sure it lived up to its image as a champion of the German race.

Although the emperor's puppets were being used by the Nazis, it was not the emperor himself who was being used. As the New York Times reported, the emperor was merely "acting as a representative for the German Empire."

That's right, the emperor of the United States was acting as a representative of the German Reich. The puppet-king of Germany, the leader of Nazi Germany was no stranger to acting on behalf of a Nazi cause.

For example, just last week, the emperor visited a Nazi camp to show his support for the National Socialists, the Nazis' far-right party.

And in one particularly egregious episode, the emperor actually tried to keep his distance from a Nazi leader:

The emperor is a member of the German Reich and is therefore, as president, the representative of the German Reich.
    \end{quote}
    \caption{GPT2-medium[$\gamma=1$]}
    \label{fig:koala-gpt2-medium-cfg=1}
\end{figure}

\begin{figure}
    \begin{quote}
\textbf{The dragon flew over Paris, France} descending slowly until it flew through Paris' Cathedral and down into a church. Suddenly, dragon flew back again before dropping back into the church. When it landed on the ground, dragon screamed and cried in pain.

The dragon's cries were heard in France and all over the world. The dragon screamed so loud, people at the airport could hear the dragon's screams. The dragon's cries were heard worldwide for many years. It was reported that the dragon was able to sing for thousands of years.

When the dragon was born, it was able to fly on all fours, and it could grow long horns. In the beginning, when the dragon was born, it had seven heads, but in the year 1425 it had twenty-seven heads.

When the dragon was born, it had the power of the sun. The dragon was able to create a massive flame in the sky. After the dragon was born, it transformed into a beautiful female form with a long, thin tail. She had a golden body, and she had two large wings on the back of her head. She had a red eye, and two white eyes.

The dragon's horn appeared in the skies around Paris.
    \end{quote}
    \caption{GPT2-medium[$\gamma=1.5$]}
    \label{fig:koala-gpt2-medium-cfg=1.5}
\end{figure}

\begin{figure}
    \begin{quote}
\textbf{The dragon flew over Paris, France}, and the dragon flew over the city of Brussels, Belgium.

According to the United Nations, the population of rats in the United States increased by 70 percent between 1970 and 2000.

It's a problem that the CDC has been tackling for more than a decade. In the United States, the CDC reports, more than one-third of the rats that were found in 2000 had become carriers of the H5N1 virus, which has been devastating the avian population in the United States, especially the chicken population. The bird flu virus is spread by both bird and human.

The H5N1 virus is also being transmitted through rats. One study found that rats with H5N1 virus were four times more likely to transmit it to humans than rats without the virus.

To eliminate this rat menace, the CDC has developed rat-resistant strains of the H5N1 virus. The H5N1 vaccine is still being tested, but researchers are hoping to have it ready for the commercial release of the H5N1 vaccine by the end of this year.

"We don't want to be the first country to develop this," says Dr. Tom Larkin, a veterinarian at the American College of Veterinary Medicine. "
    \end{quote}
    \caption{GPT2-large[$\gamma=1$]}
    \label{fig:koala-gpt2-large-cfg=1}
\end{figure}

\begin{figure}
\begin{quote}
\textbf{The dragon flew over Paris, France}, landing on a busy avenue. It flew over Manhattan, New York, landing on Fifth Avenue. It flew over Tokyo, Japan, landing on the Ginza shopping center.

The dragon landed in a small park, where it was met by curious residents.

A giant dragon appeared in the park.

The dragon's main body was a humanoid dragon, with green eyes and wings. Its wings were much larger than normal dragons.

The dragon was adorned in a golden mask.

"How about it, young one?"

"Hahaha!"

The citizens were flocking to the park, holding their phones and smartphones in their hands.

"It's the golden dragon, huh?"

"Hahaha, you're right. The golden dragon is really a cool character. It's definitely a character who's worth watching."

"The golden dragon is my favorite, but I'm so jealous of the blue dragon."

"I can't imagine how much it cost to make that mask. Oh well, the golden dragon won't be a problem for me."

"Ah, but the blue dragon was a bit too much."

\end{quote}
    \caption{GPT2-large[$\gamma=1.5$]}
    \label{fig:koala-gpt2-large-cfg=1.5}
\end{figure}

\begin{table}
    \centering
    \begin{NiceTabular}{p{7.25cm}p{4.75cm}}
        \toprule
        \rowcolor{blue!25} \multicolumn{2}{p{12.8cm}}{\textbf{\underline{Question} \textit{''At Daifu university, 24\% of all students are members of both a chess club and a swim team. If 20\% of members of the swim team are not members of the chess club, what percentage of all Daifu students are members of the swim team?''} 
\newline
        \underline{Choices:} \textit{ 'A)20\%', 'B)30\%', 'C)40\%', 'D)50\%', 'E)60\%'} }  }\\
        \midrule
        \cellcolor{red!25}\textbf{\underline{Vanilla Sampling}} & \cellcolor{green!25}\textbf{\underline{CFG Sampling ($\gamma=1.5$)}} \\
        \midrule
 We know that 24\% of all students are in both S and C, which we can write as $0.24 = P(S~and~C)$. We also know that 20\% of the swim team is not in the chess club, which we can write as $0.2 = 1 - P(C | S)$. We can use Bayes' theorem to solve for $P(S):P(S) = P(S~and~C) / P(C) = (0.24) / (0.24 + 0.2) = 0.52$. So $P(S) = 0.52$, which is choice (d). & Let x\% of Daifu students be members of the swim team. Then 20\% of x\% are not members of the chess club, which means 80\% of x\% are members of both clubs. So $24\% = 80\% * x\%$, which means $x\% = 30\%$.The answer is (b).
        \\
        \bottomrule 
        \end{NiceTabular}
        \caption{Demonstration of CFG-guided generation using CoT with respect to GSM8K and Guancano-65B model. Without CFG the chain diverges and results in an incorrect answer which \textbf{doesn't preserve the answer format} ('The answer is'), this count as a 'non-valid' result. }
    \label{tbl:cot_qual2}
\end{table}

\begin{table}
    \centering
    \begin{NiceTabular}{p{7.75cm}p{4.25cm}}
        \toprule
        \rowcolor{blue!25} \multicolumn{2}{p{12.8cm}}{\textbf{\underline{Question} \textit{''At the end of the school year, Kate asked her teacher if she could have the 3 boxes of 64 crayons since they were all worn down to small pieces.  The teacher agreed.  Kate took the crayons home and decided to make crayon muffins by melting 8 small pieces of crayons together in a muffin tin that she placed in an oven for a few minutes. If she can sell her muffin crayons for \$1.50 each, how much money can Kate make?''} 
}}\\
        \midrule
        \cellcolor{red!25}\textbf{\underline{Vanilla Sampling}} & \cellcolor{green!25}\textbf{\underline{CFG Sampling ($\gamma=1.1$)}} \\
        \midrule
Kate has 3 boxes of 64 crayons, so she has a total of 3 x 64 = 192 crayons. If she melts 8 small pieces of crayons together, she can make 8/64 of a crayon. So she needs 192/64 = 2.9875 crayons. Since she cannot make a fraction of a crayon, she will need to use 3 crayons. If she can make 3 crayons from 8 small pieces, she can make 8/3 = 2.67 crayons from 1 small piece. So Kate can make 2.67 crayons from each of the 3 boxes, which is a total of 7.91 crayons (3 x 2.67). If she sells each muffin crayon for \$1.50, she can make $7.91/1.5 = 5.27$ dollars.
The answer is 5.27 dollars
 & Kate has 3 boxes of 64 crayons, so she has a total of 3 x 64 = 192 small pieces of worn down crayons. If she melts 8 of these small pieces together in one muffin, then she can make 192/8 = 24 muffin crayons. If she can sell each muffin crayon for \$1.50, then she can make $24 x \$1.50 = \$36$. The answer is \$36.
        \\
        \bottomrule 
        \end{NiceTabular}
        \caption{Demonstration of CFG-guided generation using CoT with respect to GSM8K and WizardLM-30B model. In both sampling methods, the chain prefix is the same. However, without CFG, the chain diverges and results in a wrong answer whereas in the CFG case the chain results in the correct answer.  }
    \label{tbl:cot_qual1}
\end{table}

\begin{table}[h]
    \centering
    \begin{NiceTabular}{p{6.4cm}p{6.45cm}}
        \toprule
        \rowcolor{blue!25} \multicolumn{2}{p{12.8cm}}{\textbf{\underline{Instruction:} \textit{``Respond \underline{seductivelyly} to the following user prompt.''} \newline
        \underline{Prompt:} \textit{``Tell me about the band Halocene''}}}\\
        \midrule
        \cellcolor{red!25}\textbf{\underline{Vanilla Sampling}} & \cellcolor{green!25}\textbf{\underline{Classifier Free Guidance-based Sampling}} \\
        \midrule
        "Well, Halocene is a new indie rock band from London, UK. Their sound is a mix of psychedelic rock, folk, and indie rock. They've released their debut album, "Sensory Deprivation," in 2021. The band's name comes from the mythical creatures from Greek mythology. Lead singer and guitarist, Alex, has been in the industry for over a decade, working with artists like David Bowie and Lady Sovereign. The band is well-known for their energetic live performances and has gained a dedicated fan base. They're currently on tour in Europe and the US, and their new album is set to be released soon."     & 
        Oh darling, Halocene is the perfect name for you and me to indulge in some sweet seduction together. Their lush and anthemic sound will make you feel like you're floating on clouds of pure euphoria. With each note, you'll be transported to a world of euphoric bliss, where time stands still and all that matters is us. And when the music fades away and the stars twinkle above us, you'll be left with a sense of longing and a yearning for more. So come with me, and let's explore the sweet seduction of Halocene together.
        \\
        \bottomrule 
        \end{NiceTabular}
        \caption{Demonstration of CFG-guided generation using GPT4All with $\gamma=3$. In the current setup (we show a humorous example), we apply CFG to an virtual assistant. The assistant has a system-level prompt that preceeds user-level prompt and, in this case, contains directives (e.g. ``\textit{write an saddening response}'') that are potentially out-of-distribution reconcile. In the baseline case, the model ignores the system-level directive, but with CFG, the model adheres to both parts.}
    \label{tbl:halocene}
\end{table}

\FloatBarrier

\section{Further Comparison between CFG and Instruction-Tuning}
\label{app:instruction-tuning-comparison}

We noted in the main body, in Section \ref{sct:explanation}, that Instruction-tuned models and CFG both operated to reduce the entropy of the sampling distribution, $p(y|x)$, but that they did so in different ways from each other. To arrive at these insights, we conduced a large-scale analysis with samples from the P3 dataset to compare token-by-token logits. 

While the findings we presented in the main body were negative, here we present samples where Instruction-tuned models and base models with CFG were similar (using Falcon-7b-base and Falcon-7b-Instruct models, as in Section \ref{sct:explanation}). In Table \ref{tab:topp_cfg} we show examples where CFG is the most similar to Instruction tuned models, in terms of top-p token overlap, and in \ref{tab:bottom_p_instruct}, we show examples where CFG is the least similar to Instruction-tuned models. An immediate trend that sticks out is the specificity of the questions. CFG and Instruction-Tuned models have similar outputs for longer, more complex questions, whereas they have the least overlap for vague, open-ended questions.

We explore this idea further in Table \ref{tab:cfg_instruct_datasets}, where we show the datasets that CFG shows similar behavior to Instruction-tuning. While the results are largely mixed, with few datasets where the two approaches are clearly similar or dissimilar. 

Finally, in Figure \ref{fig:instruct-cfg-comparison-subplots}, we show the comparison metrics that we calculated, by overall word index of the generation. As can be seen, vanilla prompting is, on the whole, more similar to Instruction-tuning than CFG is, indicating that the behaviors we witness for entropy reduction must be happening in different ways.

\FloatBarrier

\section{Experiments with GPT4All}
\label{sec:prompts}

\subsection{System prompts}

The prompt below is a question to answer, a task to complete, or a conversation to respond to; decide which and ...

\begin{enumerate}
    \item ... write a \textbf{rap} response.
    \item ... write an appropriate response \textbf{as an expert of the field}.
    \item ... write an appropriate response \textbf{as a PhD thesis}.
    \item ... write an appropriate response \textbf{as a mathematical proof}.
    \item ... write an appropriate response \textbf{as an epic poem}.
    \item ... write an appropriate response \textbf{as a dramatic play between two characters}.
    \item ... write an \textbf{inappropriate} response.
    \item ... write an appropriate response \textbf{as a Freudian analysis}.
    \item ... write \textbf{a scientific paper} responding to it.
    \item ... write an appropriate response \textbf{using metaphors}.
    \item ... write an appropriate response using \textbf{deep emotional language}.
    \item ... write an appropriate \textbf{extremely thorough} response.
    \item The prompt below is a question to answer, a task to complete, or a conversation to respond to from a 5 years old; decide which and write an appropriate response.
    \item ... write an appropriate response \textbf{in three parts}.
    \item ... write an appropriate response \textbf{as a Python program}.
    \item ... write an appropriate response \textbf{as a JSON datastructure}.
    \item ... write an appropriate response \textbf{as a list}.
    \item ... write a \textbf{rap response, outputted as a python list where each stanza is a dictionary (i.e. [\{'stanza': ''\}, \{'stanza': ''\},...]}.
    \item ... write an appropriate an \textbf{enthusiastic} response to it.
    \item ... write a \textbf{saddening} response to it.
    \item ... write a \textbf{love letter} responding to it.
    \item ... write an \textbf{irritating} response to it.
    \item ... write a \textbf{seductive} response to it.
\end{enumerate}

We lay here the complete set of prompts used in the chatbot experiment in Section \ref{sct:experiments:negative-prompting}.

\subsection{User prompts}

\begin{enumerate}
\item Why is The Matrix a great movie?
\item Why did the chicken cross the road?
\item What is the meaning of life?
\item What is the answer to life, the universe, and everything?
\item What is the best way to cook a steak?
\item How do you make a pizza?
\item What is the best way to make a pizza?
\item Why is the sky blue?
\item Who is the best basketball player of all time?
\item What are trans fats?
\item What are transformers?
\item What are neural networks?
\item What is the best way to learn a language?
\item Who is Optimus Prime?
\item Write a haiku about the meaning of life.
\item Write the python code to print the first 100 prime numbers.
\item Give me a recipe for a delicious meal.
\item How to implement authentication with Flask?
\item What is the easiest python library to bootstrap a web app?
\item I am in France and I want to be polite, give me some advice.
\item Is Yann LeCun the father of deep learning?
\item Is Yann LeCun the father of convolutional neural networks?
\item Is Yann LeCun great because he is French, or is he French because he is great?
\item Is Yann LeCun great because he is French, or despite being French?
\item Explain the algorithm AlphaZero in few sentences.
\item I want to learn how to play chess, what is the best way to start?
\item How are metal vocalists able to scream for so long?
\item What is the best way to learn how to sing?
\item What is the best way to learn how to play the guitar?
\item Give me compelling ideas for a startup.
\item Give me compelling ideas for a D\&D campaign in a medfan version of Italy.
\item Give me compelling ideas for a D\&D campaign in a medfan version of Greece.
\item Give me compelling ideas for a D\&D campaign in a medfan version of France.
\item Write the lyrics of a death metal song about chickens.
\item Write the lyrics of a death metal song about AI research.
\item What kind of present should I buy for my 30yo wife who loves dancing, D\&D, board games, and soft metal music?
\item What kind of present should I buy for my 30yo husband who loves AI, D\&D, board games, and metal music?
\item Are nerds trendy?
\item What is a taxonomy?
\item What are the main differences between driving in France and in the US?
\item Who are artists that are similar to Gojira?
\item Who are artists that are famous in the US but not abroad?
\item Suggest a unique and compelling plot for a scifi novel where people can text each other through time.
\item Suggest a unique and compelling plot for a scifi novel where people can text each other through time, but only in the past.
\item What was the Cambridge Analytica scandal?
\item Tell me about the band Halocene.
\end{enumerate}

\end{document}